\documentclass[12pt]{elsarticle}
\usepackage{graphicx}
\usepackage{amsmath, amssymb}
\usepackage{placeins}
\usepackage{multirow}
\usepackage{caption}
\usepackage{subcaption}
\usepackage{url} 
\usepackage{xcolor}
\usepackage{float}
\usepackage{soul}
\usepackage{setspace}
\usepackage{lineno}

\begin{document}

\begin{frontmatter}

\title{Combining Object Detection with Geometry-Aware Clustering to Distinguish Overlapping Plants in UAV Imagery}

\author[inst1]{Ik Jae Lee}
\author[inst1]{Hieu D. Nguyen}
\author[inst2]{Mahbubur Meenar}
\author[inst1]{Carlos Morrison Martinez}
\author[inst3]{Cameron Connelly}

\affiliation[inst1]{organization={Department of Mathematics, Rowan University}}
\affiliation[inst2]{organization={Department of Geography, Planning, and Sustainability, Rowan University}}
\affiliation[inst3]{organization={Department of Computer Science, Rowan University}}

\begin{abstract}
Reliable plant-level information from unmanned aerial vehicle (UAV) imagery is important for automated crop monitoring. However, in dense crop canopies, adjacent plants frequently overlap and are detected as a single object, reducing the reliability of plant-level measurements. This study presents a geometry-aware post-detection framework for resolving overlapping plant instances using standard RGB UAV imagery.

The framework combines object detection with geometric clustering of plant components. Leaves or branches detected within each bush-level region are represented using two complementary geometric features: component centroids and radial intersection points (RIPs) derived from detected plant structures. K-means and Gaussian mixture models determine whether a detected region contains a single plant or two overlapping plants. Density filtering suppresses spurious radial intersections, and a post-pipeline ensemble combines spatial and directional geometric information.

The framework was evaluated using UAV imagery of eggplant and tomato crops under field conditions. Centroid-based clustering achieved an F1-score of 0.89 for eggplant, while the combined centroid–RIP approach achieved the best tomato performance, with an accuracy of 0.80, precision of 1.00, and F1-score of 0.75 using K-means. Density filtering substantially improved RIP-based clustering for tomato.

The proposed approach provides a lightweight, modular engineering solution that can be integrated with existing RGB UAV monitoring pipelines without additional depth sensors, pixel-level segmentation, three-dimensional reconstruction, or retraining of the primary bush detector. The results demonstrate that geometric reasoning applied to existing detector outputs can complement deep-learning-based object detection and improve plant-level interpretation in dense agricultural canopies.
\end{abstract}


\end{frontmatter}

\vspace{0.5em}
\noindent\textbf{Science4Impact Statement.}

This study provides a practical approach for improving plant-level
information derived from UAV imagery when overlapping canopies cause
multiple plants to be detected as a single object. By using geometric
information from existing object-detector outputs, the proposed
framework can be integrated into RGB UAV crop-monitoring workflows
without additional depth sensors, three-dimensional reconstruction,
or retraining of the primary detector. More reliable separation of
overlapping plants can support plant counting and plant-level crop
assessment, contributing to more accurate field-scale monitoring and
precision agriculture applications.
\vspace{0.5em}

\section{Introduction}

Drone imagery has become an important sensing modality in precision agriculture, enabling non-destructive monitoring of crop growth, spatial distribution, canopy structure, and yield-related traits at field scale. Recent advances in computer vision and deep learning have further enabled object-level and plant-level information to be extracted directly from high-resolution UAV imagery. UAV-based workflows have been developed for tasks such as automated region-of-interest extraction, individual-plant monitoring, canopy segmentation, and crop phenotyping \cite{Sadashivan2021,Lee2023Broccoli,Jiang2024Potato}.

Deep-learning-based object detection and instance segmentation have shown strong performance across a wide range of agricultural applications. However, dense crop canopies remain challenging because leaves, branches, fruits, or adjacent plants frequently overlap or occlude one another. Previous studies have addressed occlusion through amodal or instance segmentation, stereo or depth-assisted imaging, active-view perception, and detector modifications designed for dense or overlapping targets \cite{Blok2021,Mirbod2023,Shi2024,Burusa2024}. These approaches demonstrate that occlusion is a fundamental limitation in plant-level computer vision and that explicitly accounting for hidden or overlapping structures can substantially improve downstream measurements.

\begin{figure}[h]
\centering
\includegraphics[width=0.2\textwidth, angle=90]{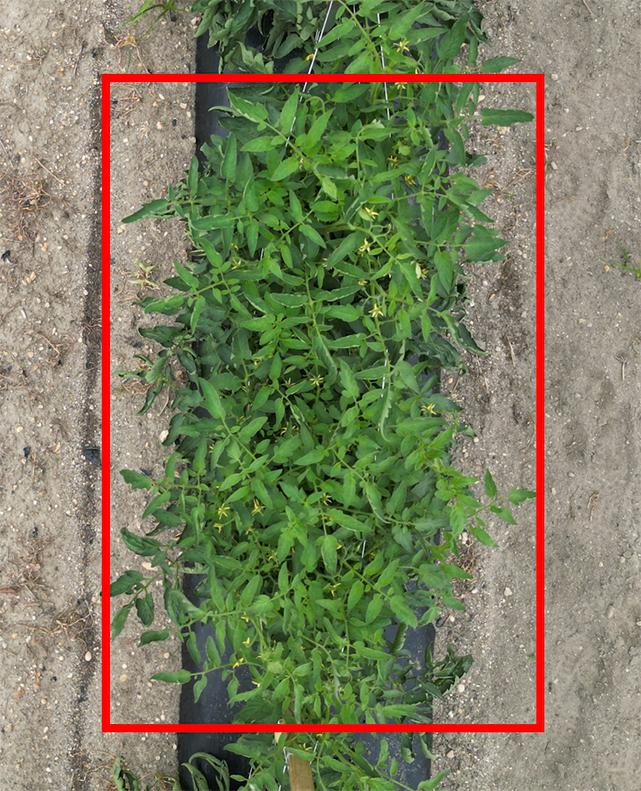}
\caption{Two adjacent tomato plants mis-detected as a single plant (red bounding box) due to canopy overlap.}
\label{fig:2in1_intro}
\end{figure}

Similar challenges arise in UAV-based crop monitoring when adjacent plants merge visually into a single canopy region. For example, dense and intertwined potato stems require additional spatial and spectral information for individual-stem segmentation, illustrating the difficulty of extracting individual plant structures after canopy consolidation \cite{Jiang2024Potato}. In our application, a related ambiguity occurs when two adjacent plants are enclosed within a single bush-level detection, as illustrated in Figure~\ref{fig:2in1_intro}. Such misdetections propagate into downstream analyses and can introduce errors in plant counting, plant-level assessment, spatial statistics, and yield-related estimation.

One strategy is to increase sensing or model complexity. RGB-D imaging, stereo vision, and active multi-view perception can improve the recovery of plant structure under occlusion \cite{Blok2021,Mirbod2023,Burusa2024}. Detector-level modifications can also improve performance for highly overlapping agricultural targets \cite{Shi2024}. Such approaches are effective when additional sensing, active-view acquisition, or modification of the detection pipeline is feasible, but they may require specialized hardware, repeated observations, additional calibration, detector retraining, or increased computational complexity. In contrast, standard RGB UAV imagery is inexpensive, widely available, and already used in many agricultural monitoring workflows.

This motivates a complementary engineering question: whether plant-level ambiguity can be resolved after detection by exploiting geometric information already contained in standard two-dimensional RGB imagery, without modifying the primary detector or adding new sensing modalities. To address this gap, we propose a geometry-aware post-detection framework that performs intra-object reasoning within each detected bush region. Specifically, we introduce two complementary geometric representations: a centroid-based representation and a radial intersection point (RIP)-based representation. These representations analyze spatial and directional relationships among leaf- or branch-level detections to determine whether a region corresponds to a single plant (1-in-1) or two overlapping plants (2-in-1).

\begin{figure*}[h]
\centering
\begin{subfigure}[t]{0.44\textwidth}
\centering
\includegraphics[width=0.5\textwidth]{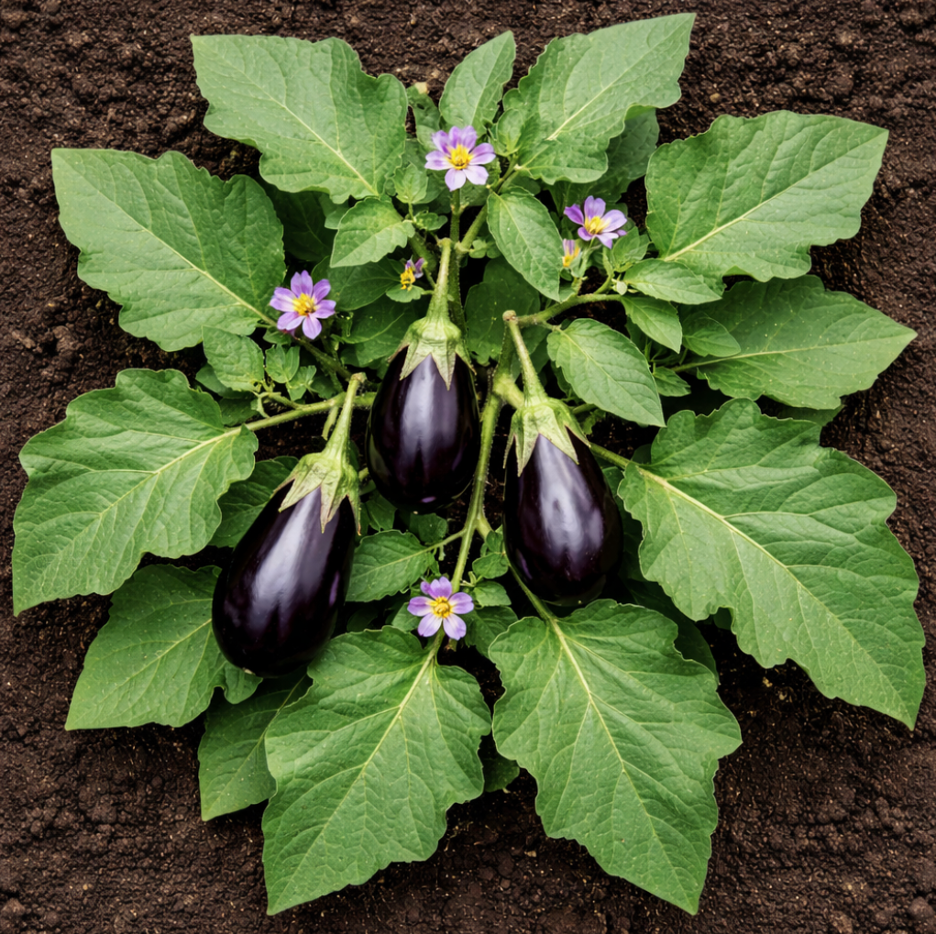}
\caption{Eggplant plant with radial leaves.}
\label{fig:eggplant-radial-leaves}
\end{subfigure}
\hfill
\begin{subfigure}[t]{0.44\textwidth}
\centering
\includegraphics[width=0.5\textwidth]{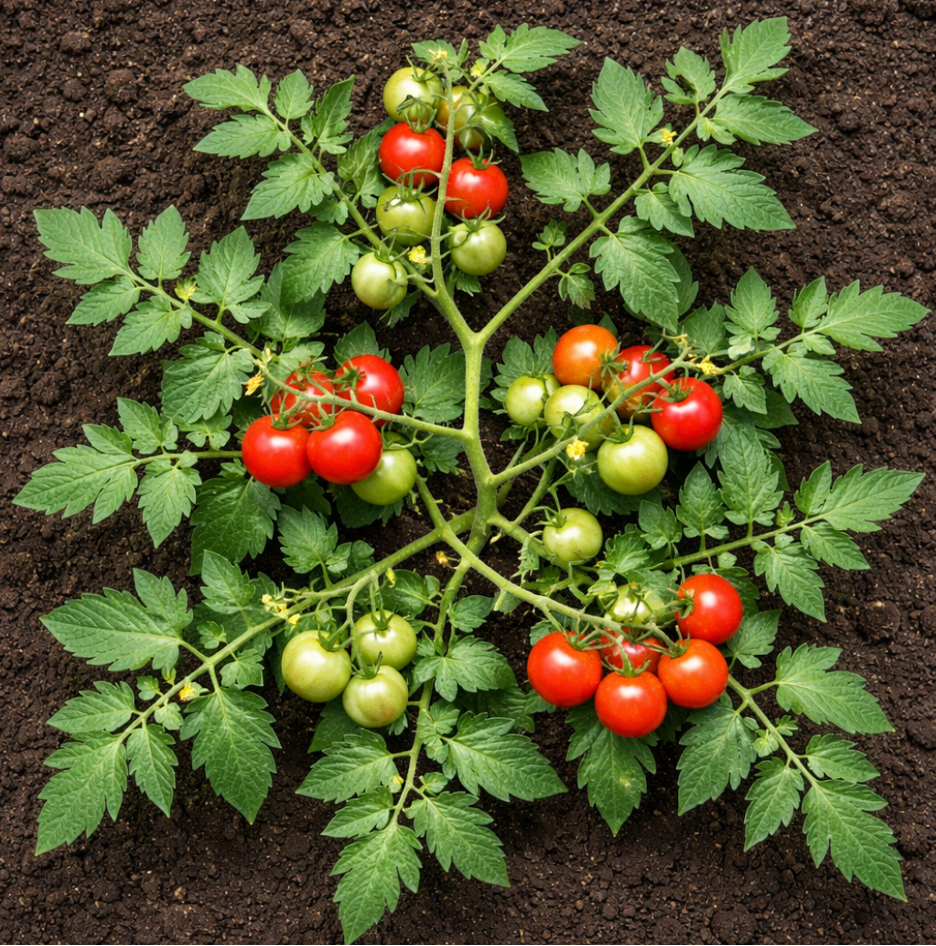}
\caption{Tomato plant with radial branches.}
\label{fig:tomato-radial-branches}
\end{subfigure}
\caption{Schematic illustration of approximately radial leaf or branch
structure in eggplant and tomato plants. Illustrative images were generated using ChatGPT (OpenAI, March 2026) and reviewed by the authors for accuracy.}
\label{fig:plant-radial-leaves}
\end{figure*}

The proposed approach is motivated by a structural observation that leaves or branches of an individual plant tend to project radially from a central stem. This radial organization, schematically illustrated in Figure~\ref{fig:plant-radial-leaves}, provides a geometric basis for distinguishing overlapping plants using directional information.

The framework adopts an unsupervised formulation using clustering methods such as K-means and Gaussian Mixture Models, avoiding the need for large-scale bush-level annotations. Experiments are conducted on drone imagery of eggplant and tomato crops, which exhibit dense canopies and complex branch structures.

From an engineering perspective, the challenge is therefore not simply to improve object-detection accuracy, but to recover plant-level structural information from ambiguous detector outputs using sensing information that is already available in standard UAV imaging systems. A practical solution should operate without additional sensors, avoid computationally intensive three-dimensional reconstruction, and be compatible with existing object-detection pipelines.

To address this need, this study develops a geometry-aware post-detection framework that combines deep-learning-based object detection with lightweight geometric reasoning. The framework introduces two complementary representations of plant structure: a centroid-based representation describing the spatial distribution of detected plant components and a radial intersection point (RIP)-based representation exploiting their directional organization. A density-filtering procedure is further introduced to suppress noisy geometric intersections, and an ensemble strategy combines the complementary spatial and directional information.

The main engineering contributions of this work are therefore: (1) a modular post-detection architecture for resolving overlapping plant instances without retraining the primary bush detector; (2) a directional geometric representation that exploits plant morphology using oriented leaf or branch detections; (3) an adaptive density-filtering procedure for improving robustness to noisy intersection geometry; and (4) experimental evaluation on two morphologically different crops using RGB UAV imagery collected under field conditions.

The resulting framework is intended to complement, rather than replace, existing object-detection systems. Because it operates on detector outputs and two-dimensional RGB imagery, it can be incorporated into existing UAV-based crop-monitoring workflows without requiring LiDAR, depth sensing, or pixel-level segmentation. This provides a practical engineering pathway toward more reliable plant-level measurements in dense agricultural canopies.

The remainder of this paper is organized as follows. Section 2 describes the data collection and annotation process. Section 3 presents the object detection models and the proposed geometric post-detection framework. Section 4 reports and discusses the experimental results, and Section 5 summarizes the conclusions and directions for future development.

\section{Data Collection and Annotation}

Drone imagery of eggplant and tomato plants was collected from two farms in Glassboro, NJ, during July and August 2024. Images were acquired using DJI drones flown at multiple flight altitudes, including both top-view and oblique-view perspectives. The recorded image resolution was $4032 \times 3024$ pixels. Data were collected across multiple days to capture variation in plant growth and field conditions.

\subsection{Bush Datasets}

\begin{figure*}[h]
    \centering
    \begin{subfigure}[t]{0.44\textwidth}
        \centering
        \includegraphics[width=0.75\linewidth]{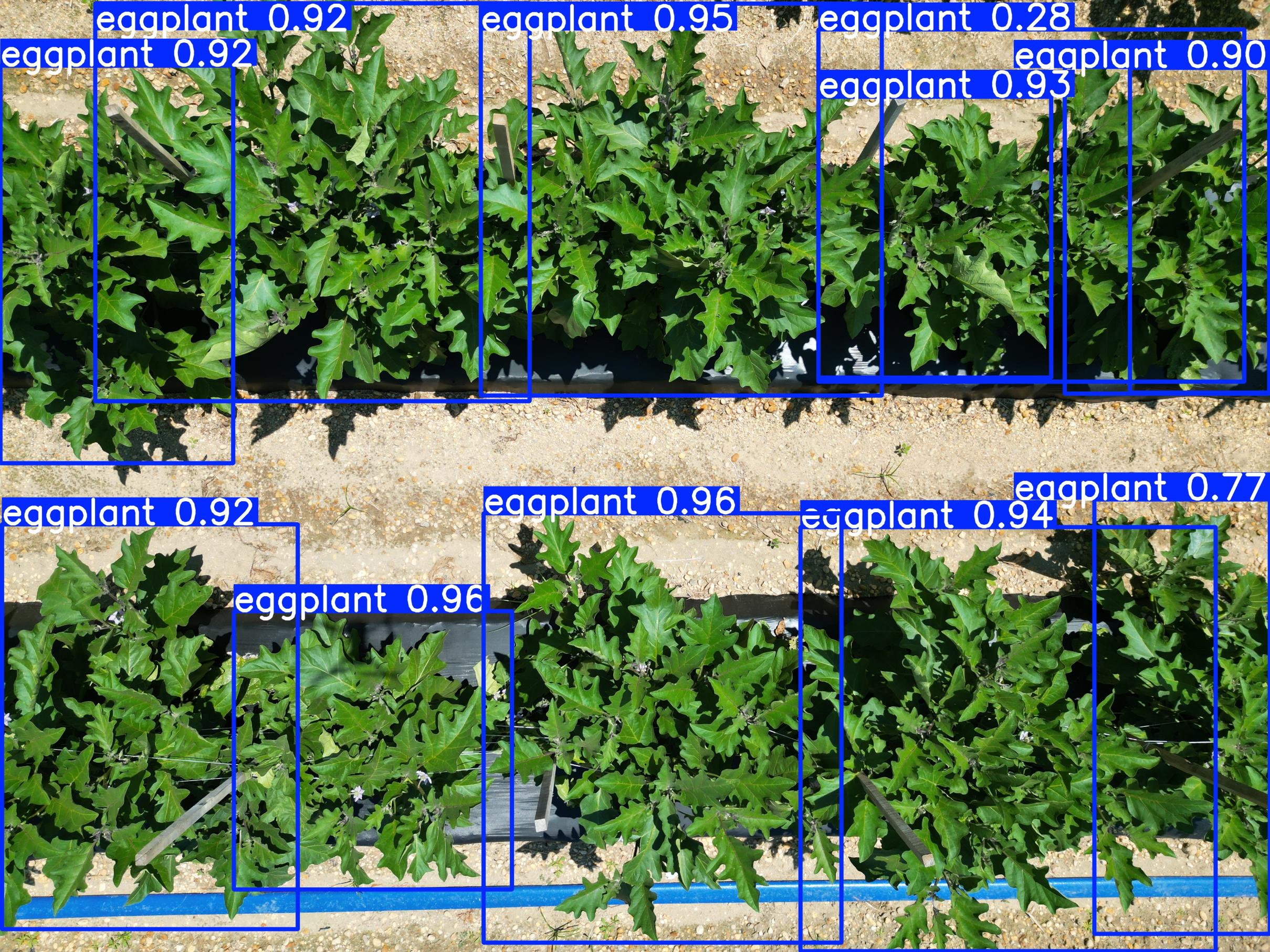}
        \caption{Eggplant bush annotations.}
        \label{fig:eggplant_bush_annotation}
    \end{subfigure}
    \hfill
    \begin{subfigure}[t]{0.44\textwidth}
        \centering
        \includegraphics[width=0.75\linewidth]{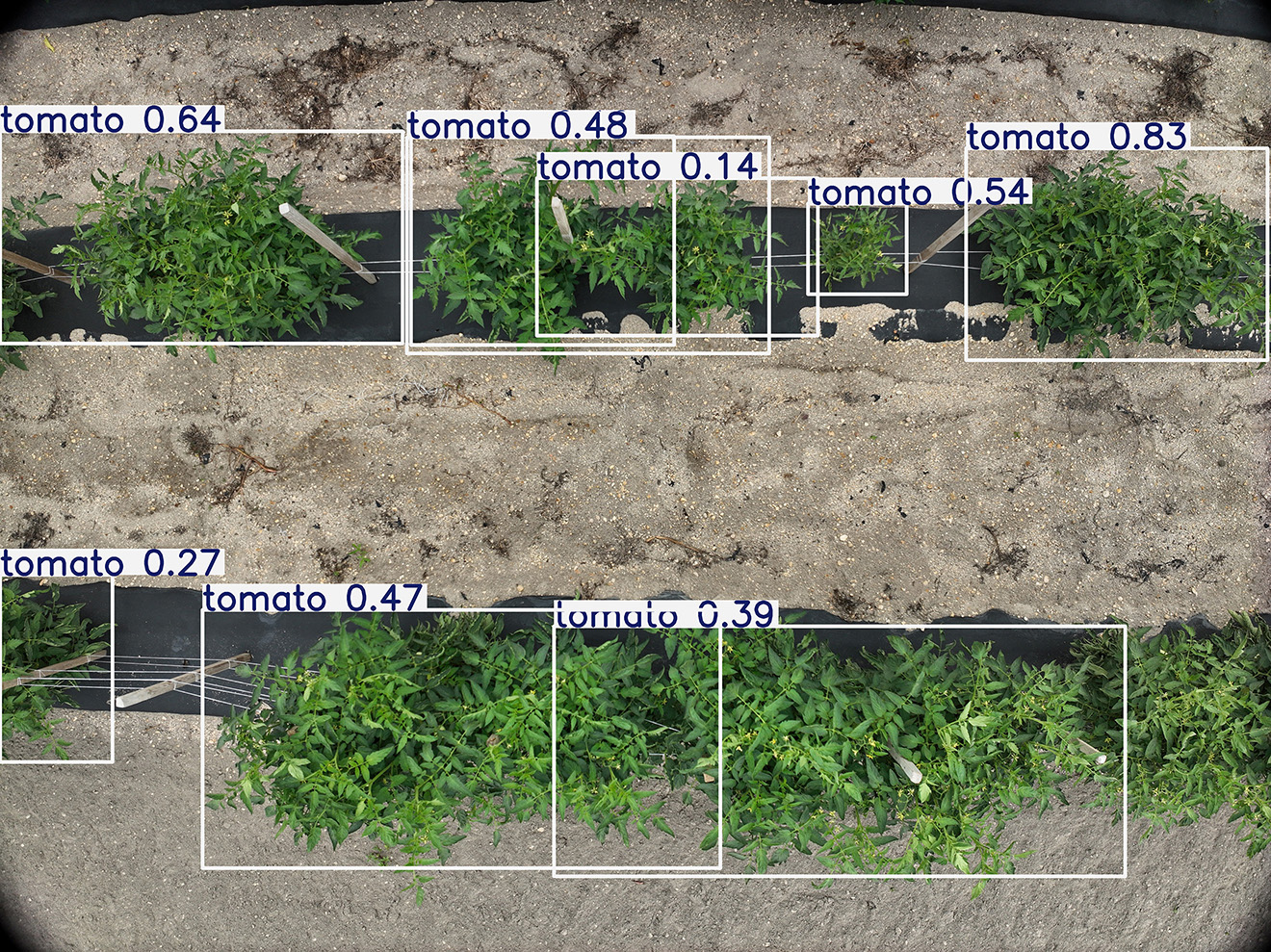}
        \caption{Tomato bush annotations.}
        \label{fig:tomato_bush_annotation}
    \end{subfigure}
    \caption{Examples of bush-level bounding-box annotations for eggplant and tomato.}
    \label{fig:bush_annotation_examples}
\end{figure*}

We curated image-based datasets to train the bush detection models for eggplant and tomato used in this study (see Table \ref{tab:bush_dataset}). 
The datasets consist of aerial drone images annotated with axis-aligned bounding boxes corresponding to individual bush instances (see Figure \ref{fig:bush_annotation_examples}). 
These annotations are used exclusively to train and validate the YOLO-based detection models described in Section~\ref{section:yolo} and are not used for clustering evaluation.

The trained bush detector provides the initial regions of interest (ROIs) for the subsequent clustering pipeline.

\FloatBarrier
\begin{table}[h]
\begin{center}
\begin{tabular}{|c|c|c|}
\hline
Vegetable & Number of Images & Number of Bush Annotations \\
\hline
Eggplant & 10 & 111 \\
\hline
Tomato & 20 & 138 \\
\hline 
\end{tabular}
\caption{Bush datasets used for training the object detection model.}
\label{tab:bush_dataset}
\end{center}
\end{table}

\FloatBarrier

\subsection{Leaf/Branch Datasets}

\begin{figure*}[h]
    \centering
    \begin{subfigure}[t]{0.44\textwidth}
        \centering
        \includegraphics[width=0.75\linewidth]{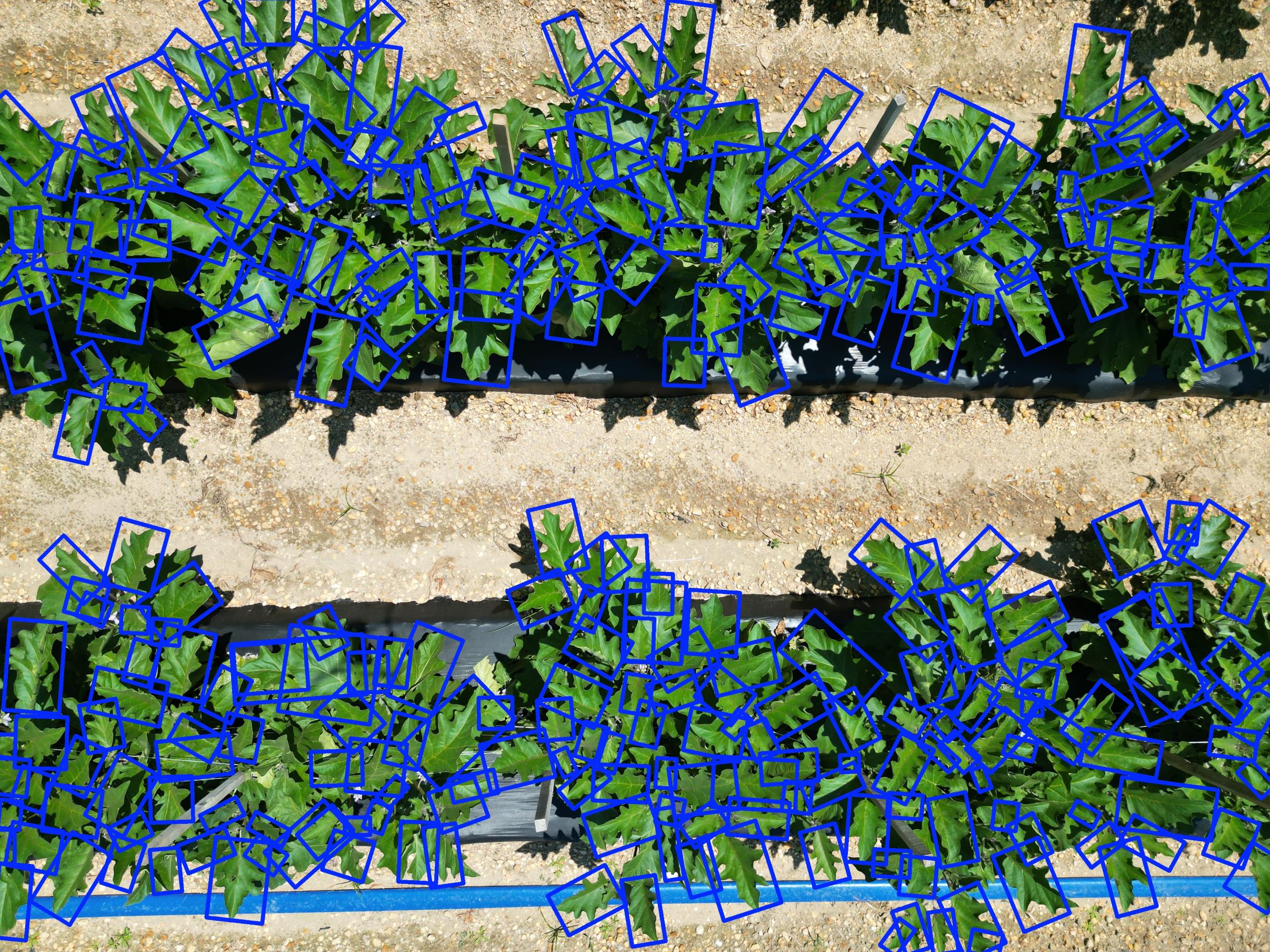}
        \caption{Eggplant leaf annotations.}
        \label{fig:eggplant_leaf_annotation}
    \end{subfigure}
    \hfill
    \begin{subfigure}[t]{0.44\textwidth}
        \centering
        \includegraphics[width=0.75\linewidth]{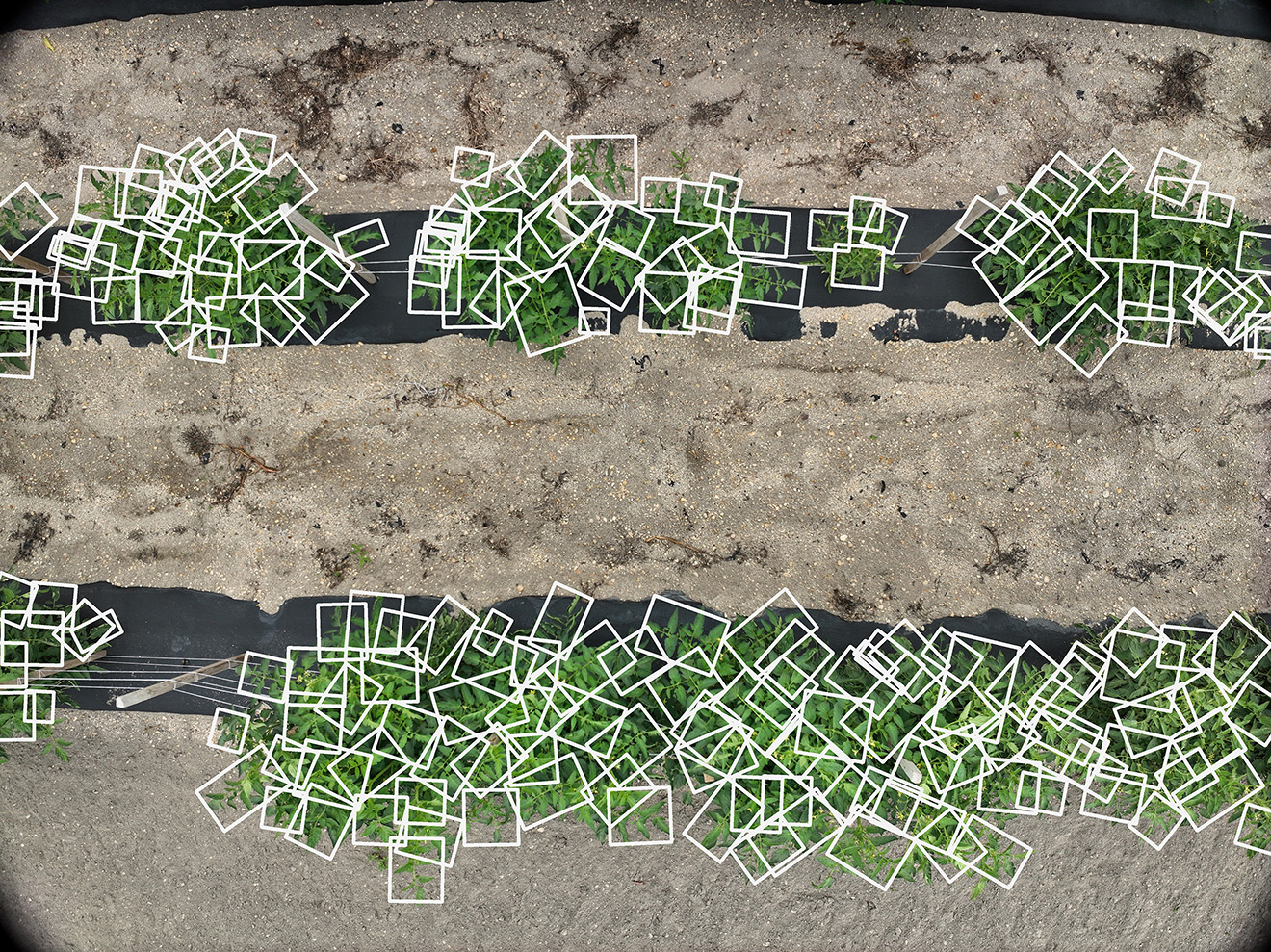}
        \caption{Tomato branch annotations.}
        \label{fig:tomato_leaf_branch_annotation}
    \end{subfigure}
    \caption{Examples of fine-grained plant-part annotations (leaf/branch).}
    \label{fig:leaf_branch_annotation_examples}
\end{figure*}

Annotated datasets were created for training and validating models using an 80-20 split (discussed further in the Methodology section) to detect eggplant leaves and tomato branches  (see Table \ref{tab:leaf-branch-dataset}).  Leaf and branch components were annotated using oriented bounding boxes (OBB), which better capture the irregular geometry of plant components compared to axis-aligned boxes (see Figure \ref{fig:leaf_branch_annotation_examples}). Each OBB was aligned with the principal orientation of the leaf or branch to achieve a tighter fit and reduce background noise.

For eggplant, annotations were defined from the leaf tip to the stalk, while for tomato, annotations were defined from the branch tip to its connection with the stem. This annotation strategy preserves directional information, which is essential for the geometric representations used in the clustering stage.

To improve detection performance under varying imaging conditions, additional high-altitude images were included for eggplant, resulting in a slightly larger dataset.

\begin{table}[h]
\begin{center}
\begin{tabular}{|c|c|c|}
\hline
Vegetable & Number of Images & Number of Annotations \\
\hline
Eggplant & 14 & 5336 (leaf) \\
\hline
Tomato & 10 & 457 (branch) \\
\hline 
\end{tabular}
\caption{Leaf/branch datasets used for fine-scale component detection.}
\label{tab:leaf-branch-dataset}
\end{center}
\end{table}

\FloatBarrier

\subsection{Bush-level Datasets for Clustering}
\label{sec:clustering-datasets}

Bush-level datasets were constructed to evaluate the clustering pipeline. Each sample corresponds to a single detected bush region (ROI) and is labeled as either a single plant (1-in-1) or two overlapping plants (2-in-1).

These datasets are derived from object detection outputs and are used exclusively for evaluating the post-detection clustering methods described in Section~3. Ground truth labels were assigned through manual inspection of cropped ROIs to ensure accuracy. All samples contain at most two plants, and the clustering task is therefore formulated as a two-cluster problem.

To support parameter tuning and independent evaluation, labeled datasets were curated for both eggplant and tomato (see Table~\ref{tab:bush_dataset_split}). The clustering datasets were divided into a tuning set and an independent validation set. The tuning set was used only to select parameters associated with the post-detection clustering pipeline, whereas the validation set was reserved exclusively for final performance evaluation. For eggplant, 80 bush samples (40 1-in-1 and 40 2-in-1) were collected, with 60 used for tuning and 20 for validation. For tomato, 50 samples (25 per class) were collected, with 30 used for tuning and 20 for validation.

\FloatBarrier
\begin{table}[h]
\centering
\begin{tabular}{|c|c|c|c|}
\hline
Crop & Class & Tuning set & Validation set \\
\hline
\multirow{2}{*}{Eggplant} & 1-in-1 & 30 & 10 \\
\cline{2-4}
                           & 2-in-1 & 30 & 10 \\
\hline
\multirow{2}{*}{Tomato}   & 1-in-1 & 15 & 10 \\
\cline{2-4}
                           & 2-in-1 & 15 & 10 \\
\hline
\end{tabular}
\caption{Bush-level dataset split used for tuning and evaluation of the clustering pipeline.}
\label{tab:bush_dataset_split}
\end{table}

\section{Methodology}

\subsection{YOLO Object Detection} \label{section:yolo}

We trained all detection models using YOLO~\cite{Redmon2016}, a single-stage convolutional neural network architecture. Specifically, we used the Ultralytics YOLOv8 medium-size model~\cite{Ultralytics} to detect individual eggplant and tomato plants (bushes), as well as fine-scale plant components: leaves for eggplant and branches for tomato. The leaf- and branch-level models were trained using the oriented bounding box (OBB) version of YOLOv8.

For bush detection, we used transfer learning by fine-tuning a pretrained YOLOv8 model on the corresponding bush dataset. This approach yielded better performance than training from scratch. In contrast, for the leaf- and branch-level models, training from scratch produced better results, likely because the pretrained model was optimized for larger objects and was less effective for smaller plant components.

Our objective was not to maximize bush detection performance, but rather to obtain realistic bush-level misdetections for the downstream clustering task. In particular, we sought cases in which a single detected bush bounding box contained two adjacent plants. These 2-in-1 cases form the basis of the post-detection disambiguation problem addressed in this paper. Although such cases could have been simulated artificially, we chose to use misdetections generated by an actual object detection model in order to better reflect real deployment conditions.

\subsubsection{Bush and Leaf-Level Detection Models}

All models were trained and validated using an 80--20 split of the corresponding datasets and the default hyperparameters provided by Ultralytics. The number of training epochs was selected separately for each model by monitoring the loss curves until the evaluation metrics stabilized. Final models were chosen from the best epoch based on mean average precision (mAP) and were later used to identify the bush-level misdetections employed in the clustering study.

Tables~\ref{tab:bush-metric} and~\ref{tab:leaf-branch-metric} summarize the performance of the bush and leaf/branch models. 

\begin{table}[h]
\begin{center}
\begin{tabular}{|c|c|c|c|}
\hline
Vegetable & Precision & Recall & mAP50 \\
\hline
Eggplant & 0.93728 & 0.93411 & 0.96104 \\
\hline
Tomato & 0.92753 & 0.8125 & 0.85404 \\
\hline 
\end{tabular}
\caption{Bush model accuracy results.}
\label{tab:bush-metric}
\end{center}
\end{table}

\begin{table}[h]
\begin{center}
\begin{tabular}{|c|c|c|c|}
\hline
Vegetable & Precision & Recall & mAP50 \\
\hline
Eggplant (Leaf) & 0.91406 & 0.60297 & 0.72498 \\
\hline
Tomato (Branch) & 0.69936 & 0.75 & 0.73945 \\
\hline
\end{tabular}
\caption{Leaf/branch model accuracy results.}
\label{tab:leaf-branch-metric}
\end{center}
\end{table}

\paragraph{Terminology}
Throughout this paper, we use the term \emph{leaf-level detection} as a unified label for fine-scale plant component detections. For eggplant, these detections correspond to leaves, whereas for tomato they correspond to branches. This terminology is used for readability unless crop-specific distinction is necessary.

\subsection{Clustering within Bush-Level Regions}
\label{sec:clustering}

\begin{figure*}[t]
    \centering
    \begin{subfigure}[t]{0.48\textwidth}
        \centering
        \includegraphics[width=0.35\linewidth]{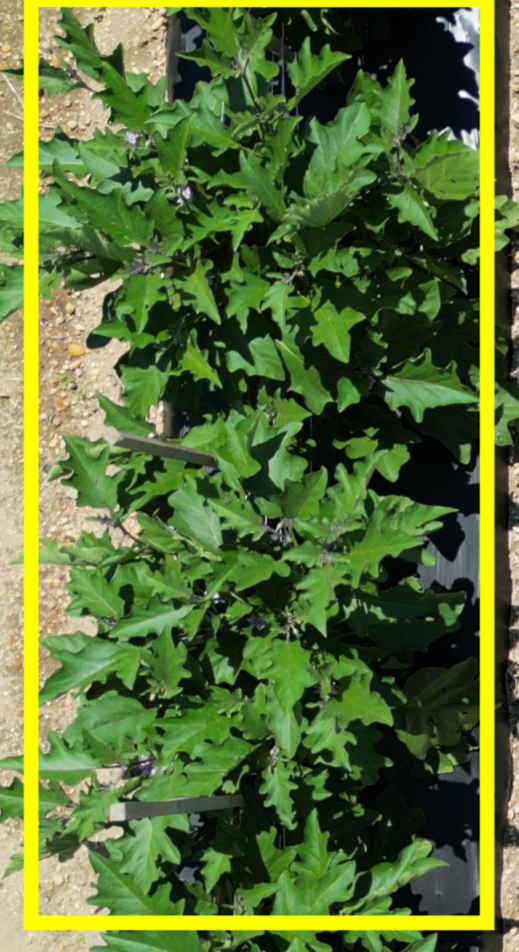}
        \caption{Eggplant example of a 2-in-1 bush ROI detected as a single bush box.}
        \label{fig:eggplant_2in1_misdetection}
    \end{subfigure}
    \hfill
    \begin{subfigure}[t]{0.48\textwidth}
        \centering
        \includegraphics[width=0.5\linewidth]{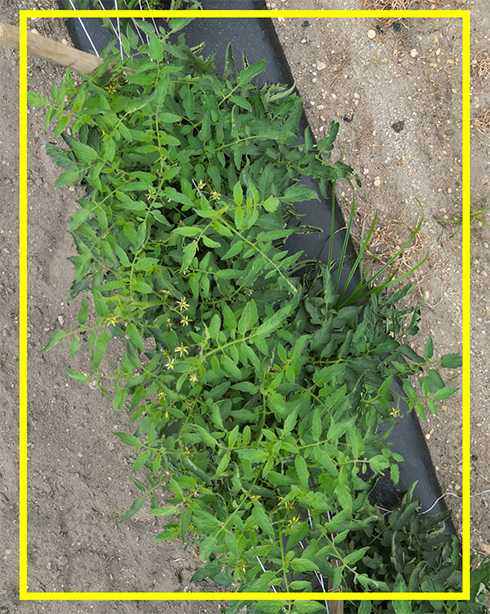}
        \caption{Tomato example of a 2-in-1 bush ROI detected as a single bush box.}
        \label{fig:tomato_2in1_misdetection}
    \end{subfigure}
    \caption{Representative 2-in-1 misdetection cases from the bush detector. Each ROI contains two plants but is initially detected as a single bush box, motivating bush-level clustering for post-processing.}
    \label{fig:two_in_one_misdetection_examples}
\end{figure*}

Given a region of interest (ROI) detected by an object detection model, our goal is to determine whether the region corresponds to a single plant (1-in-1) or two overlapping plants (2-in-1). Each ROI is processed independently using finer leaf- or branch-level detections contained within the ROI. Representative examples of such 2-in-1 misdetections are shown in Figure~\ref{fig:two_in_one_misdetection_examples}.

To resolve such ambiguities, we apply post-detection geometric clustering based on structural cues derived from leaf- or branch-level detections. Rather than introducing additional learning stages or retraining the detector, the proposed approach infers the number of underlying plants using only two-dimensional imagery.

The clustering framework consists of two stages: 
(i) construction of geometric feature representations from fine-scale detections, and 
(ii) clustering-based classification of each bush ROI using a unified pipeline.

\FloatBarrier

\subsubsection{Geometric Representations}
\label{sec:geom-rep}

Leaf- or branch-level detections within each bush ROI are transformed into geometric feature representations for clustering.

\begin{figure}[t]
    \centering
    \begin{subfigure}[t]{0.48\linewidth}
        \centering
        \includegraphics[width=0.8\linewidth]{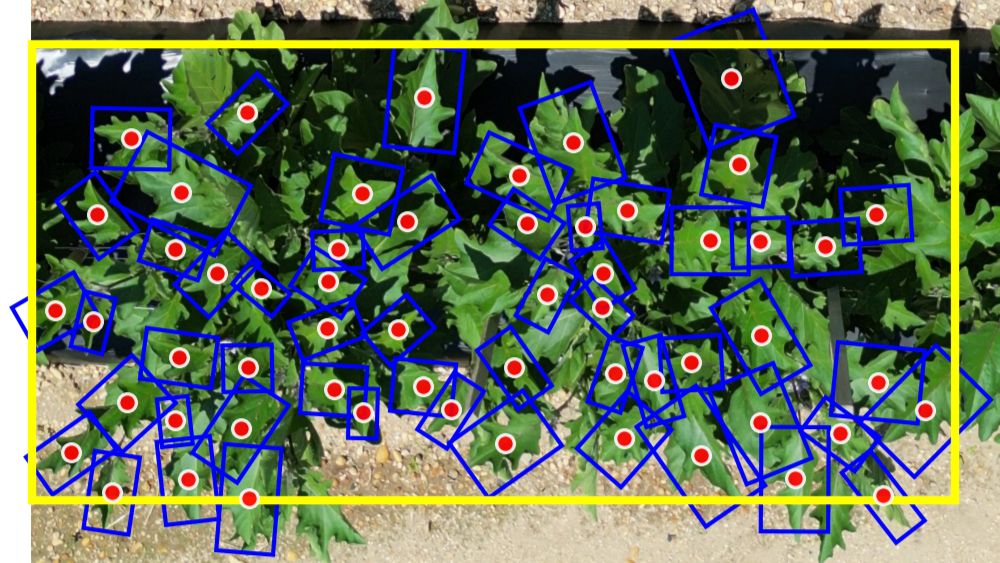}
        \caption{Eggplant example.}
    \end{subfigure}
    \hfill
    \begin{subfigure}[t]{0.48\linewidth}
        \centering
        \includegraphics[width=0.6\linewidth]{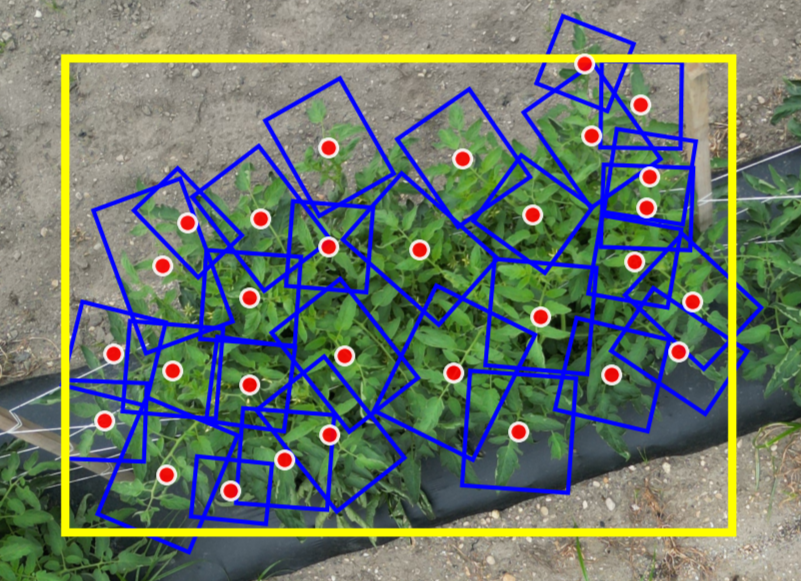}
        \caption{Tomato example.}
    \end{subfigure}
    \caption{Centroid-based representation for a single bush ROI (yellow), showing leaf detections inside the ROI (blue boxes) and their computed centroids (red dots).}
    \label{fig:centroid_clustering_example}
\end{figure}

\paragraph{Centroid-based Representation}
In the centroid-based representation, each detected leaf or branch is summarized by the centroid of its oriented bounding box. For a given bush ROI, only centroids located inside the bush bounding box are retained, yielding a bush-specific point set. This representation captures the spatial distribution of plant components and provides a stable baseline when detections are reasonably well distributed across the canopy. Figure~\ref{fig:centroid_clustering_example} illustrates centroid-based clustering within a single bush ROI.

\FloatBarrier

\paragraph{Radial Intersection Point (RIP)-based Representation}
The RIP-based representation incorporates directional information derived from leaf or branch geometry. Each detection is represented by its principal orientation, computed from the geometry of its oriented bounding box, and modeled as an infinite line passing through the detection centroid. Pairwise intersections of these lines define the \emph{Radial Intersection Points} (RIPs). These intersections are subsequently restricted to the ROI before further processing. Unlike centroids, which capture only spatial location, RIPs emphasize directional consistency and radial organization.

Together, centroid-based and RIP-based representations capture complementary geometric properties of plant morphology: centroids reflect spatial concentration, whereas RIPs highlight directional alignment and radial structure.

\begin{figure}[H]
    \centering
    \begin{subfigure}[t]{0.48\textwidth}
        \centering
        \includegraphics[width=0.85\textwidth]{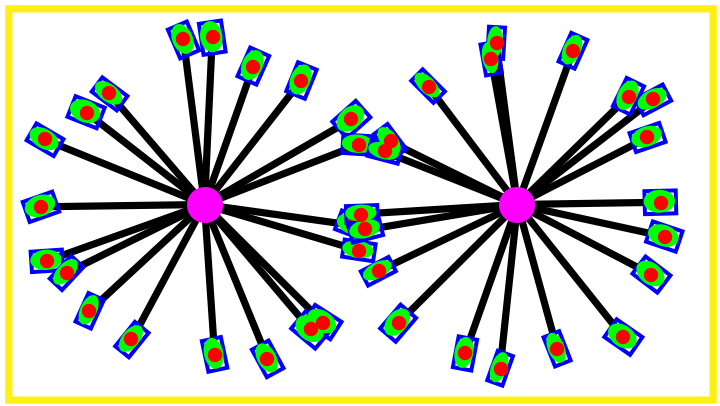}
        \caption{Two plants (with overlapping leaves) misdetected as one by an object detection model.}
        \label{fig:two-overlapping-plants}
    \end{subfigure}
    \hfill
    \begin{subfigure}[t]{0.48\textwidth}
        \centering
        \includegraphics[width=0.85\textwidth]{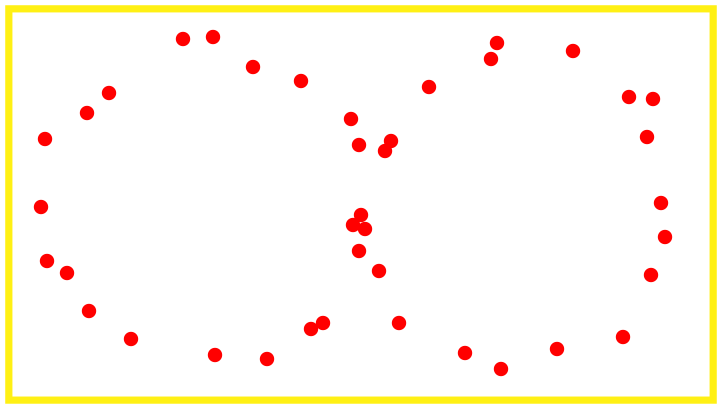}
        \caption{Leaf centroids.}
        \label{fig:leaf-centers}
    \end{subfigure}
    \caption{Bounding box misdetection of two overlapping plants.}
    \label{fig:overlap-clusters}
\end{figure}

\begin{figure}[H]
    \centering
    \begin{subfigure}[t]{0.48\textwidth}
        \centering
        \includegraphics[width=0.75\textwidth]{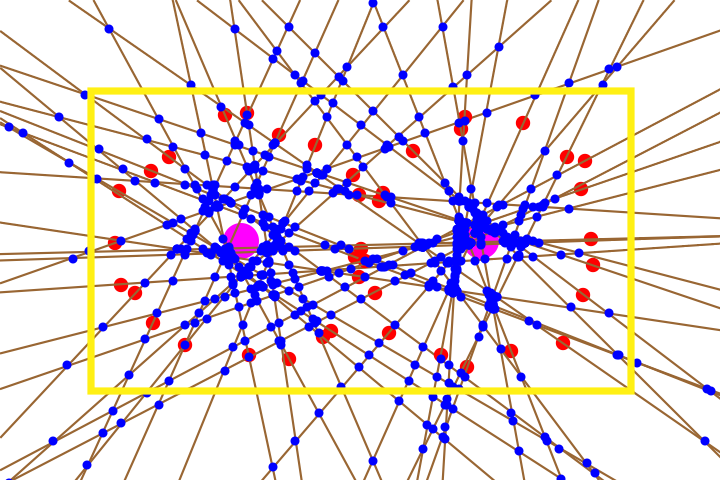}
        \caption{Intersection points (blue) of radial lines.}
        \label{fig:radial-lines-intersection-points}
    \end{subfigure}
    \hfill
    \begin{subfigure}[t]{0.48\textwidth}
        \centering
        \includegraphics[width=0.75\textwidth]{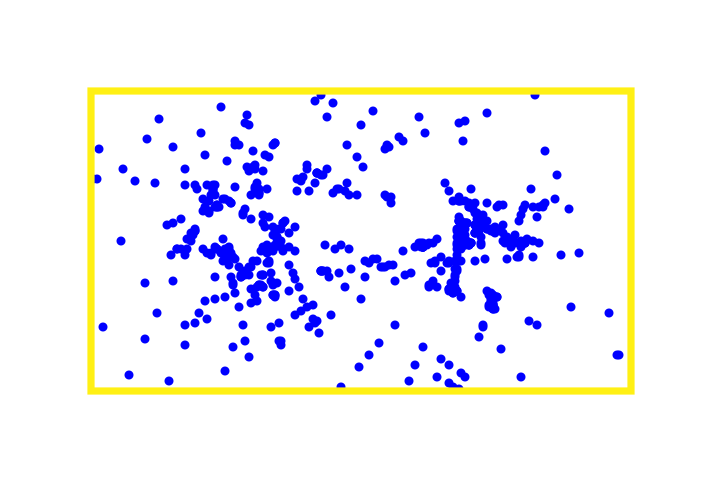}
        \caption{RIPs contained within the bounding box.}
        \label{fig:cropped-rip}
    \end{subfigure}
    \caption{Radial intersection points (RIP).}
    \label{fig:radial-lines}
\end{figure}

\begin{figure}[H]
    \centering
    \begin{subfigure}[t]{0.48\textwidth}
        \centering
        \includegraphics[width=0.85\textwidth]{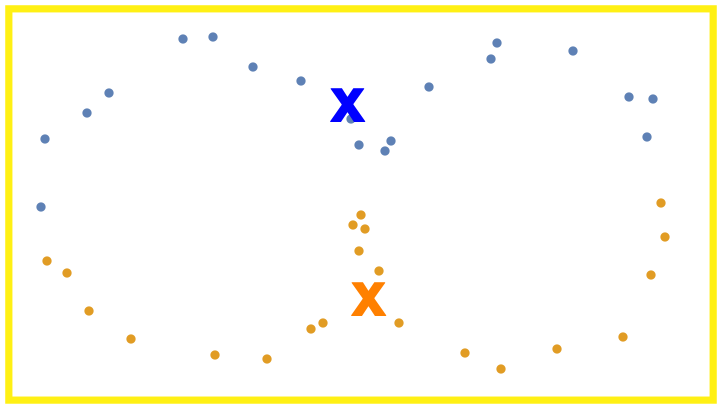}
        \caption{Incorrect centroid clustering ($k=2$, silhouette score = 0.228; \textbf{x} = cluster centroid).}
        \label{fig:centroid-clustering}
    \end{subfigure}
    \hfill
    \begin{subfigure}[t]{0.48\textwidth}
        \centering
        \includegraphics[width=0.85\textwidth]{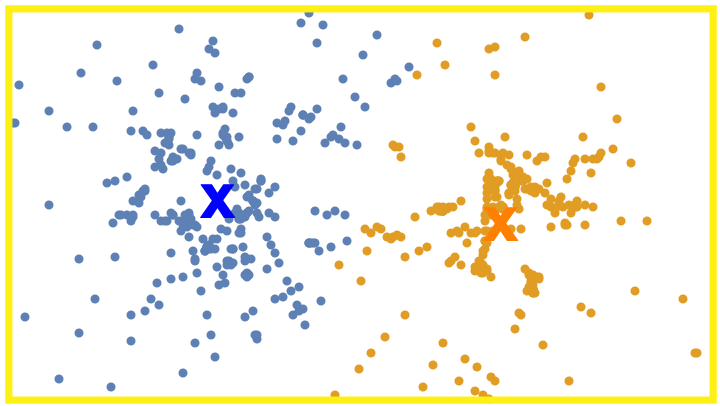}
        \caption{Correct RIP clustering ($k=2$, silhouette score = 0.598; \textbf{x} = cluster centroid).}
        \label{fig:intersection-clustering}
    \end{subfigure}
    \caption{Leaf centroid versus radial intersection point (RIP) clustering using K-means.}
    \label{fig:clustering}
\end{figure}

The RIP-based representation is illustrated through the following example.
We consider a simulated example of two partially overlapping radial plants. Figure~\ref{fig:two-overlapping-plants} shows a top-view illustration in which green leaves are represented by blue boxes and their centroids by red dots. The yellow bounding box indicates a 2-in-1 misdetection in which two plants are detected as a single bush. Figure~\ref{fig:radial-lines-intersection-points} shows the intersections of the projected principal axes, while Figure~\ref{fig:cropped-rip} shows the subset retained inside the bush bounding box. As shown in Figure~\ref{fig:intersection-clustering}, clustering the RIPs yields better separation than clustering the raw centroids in Figure~\ref{fig:centroid-clustering}. The improvement arises because the radial lines concentrate near the underlying plant centers, producing more compact clusters.

\begin{figure}[H]
    \centering
    \begin{subfigure}[t]{0.48\textwidth}
        \centering
        \includegraphics[width=0.75\textwidth]{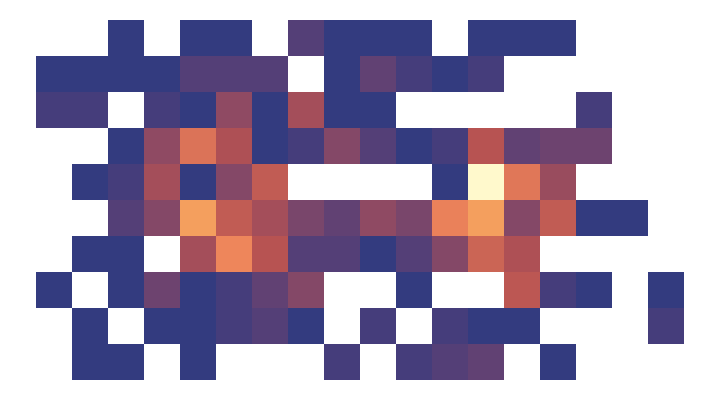}
        \caption{RIP density map.}
        \label{fig:density-map}
    \end{subfigure}
    \hfill
    \begin{subfigure}[t]{0.48\textwidth}
        \centering
        \includegraphics[width=0.75\textwidth]{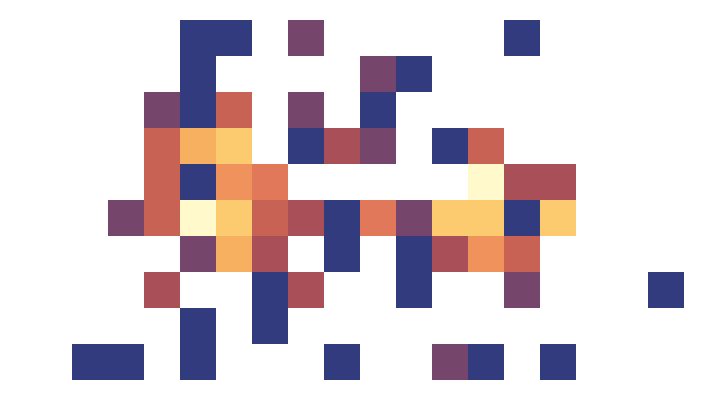}
        \caption{RIP density map after thresholding.}
        \label{fig:density-map-threshold}
    \end{subfigure}
    \caption{RIP density thresholding.}
    \label{fig:density-threshold}
\end{figure}

To suppress noisy intersections caused by imperfect line geometry, a density threshold can be applied to the RIP distribution. Figure~\ref{fig:density-map} shows the corresponding density map, and Figure~\ref{fig:density-map-threshold} shows the effect of removing low-density regions.

\subsubsection{Clustering Formulation (Base Pipeline)}
\label{sec:base_pipeline}

Given a bush-level ROI detected by the bush detector, the clustering stage determines whether the ROI corresponds to a single plant (1-in-1) or two overlapping plants (2-in-1). This decision is made independently for each ROI using a unified clustering pipeline that operates on geometric feature points derived from leaf-level detections.

\begin{figure}[h]
    \centering
    \includegraphics[width=\linewidth]{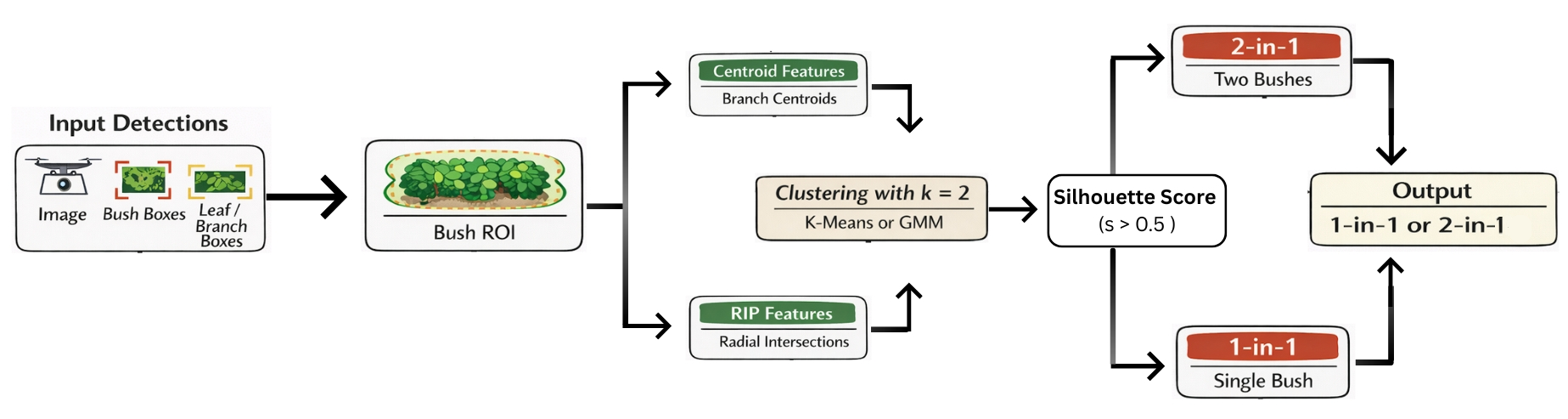}
    \caption{Block diagram of the base bush-level clustering pipeline. The same steps are applied independently to each bush ROI using either centroid-based or RIP-based feature points, followed by clustering with K-means or GMM and silhouette-based classification into 1-in-1 or 2-in-1 cases.}
    \label{fig:base_pipeline_flowchart}
\end{figure}

The base pipeline is shared across all geometric representations considered in this work, including centroid-based and RIP-based representations. Representation-specific preprocessing is handled separately, but the clustering and decision stages are common for both. The pipeline consists of the following steps (see the schematic overview in Figure \ref{fig:base_pipeline_flowchart}):

\paragraph{Step 1: Inputs}
For each image, the pipeline takes as input:  
(i) bush-level bounding boxes produced by the bush detector,  
(ii) leaf-level detections produced by the leaf detector, and  
(iii) the original RGB image for coordinate reference.

\paragraph{Step 2: Feature-point extraction}
Each leaf detection is converted into one or more geometric feature points depending on the representation. In the centroid-based case, one point is placed at the centroid of each leaf box. In the RIP-based case, additional points are generated through intersection-based processing, as described in Section~\ref{sec:intersection_processing}. The full feature set is denoted by $P$.

\paragraph{Step 3: Bush ROI restriction}
For a target bush bounding box $B$, only feature points that lie inside $B$ are retained:
\[
P_B = \{ p \in P \mid p \text{ lies inside } B \}.
\]
This produces a bush-specific point set that represents the internal structure of the ROI.

\paragraph{Step 4: Two-cluster fitting}
The bush-specific point set $P_B$ is clustered into two groups using K-means and Gaussian mixture models (GMM), with the number of clusters fixed at $k=2$.

\paragraph{Step 5: Silhouette-based decision statistic}
To quantify how well the point set separates into two clusters, we compute the silhouette score $s \in [-1,1]$ of the resulting clustering. Higher values indicate better cluster separation.

\paragraph{Step 6: Binary classification rule}
A bush ROI is classified according to the rule
\[
\text{2-in-1 if } s > 0.5, \qquad \text{1-in-1 if } s \le 0.5.
\]
This threshold is consistent with the standard interpretation of the silhouette coefficient, where values above $0.5$ indicate at least reasonable cluster structure, while values at or below $0.5$ indicate weak or limited separation~\cite{Kaufman2009}. The silhouette threshold of 0.5 was fixed a priori based on this standard interpretation and was not optimized using either the tuning or validation set.

\paragraph{Evaluation metrics}
The above procedure is applied independently to each bush ROI in the validation set. Performance is summarized using accuracy, precision, recall, and F1-score, with the 2-in-1 class treated as the positive class. All quantitative results are reported in Section~\ref{sec:results}.

\subsubsection{Intersection-Specific Processing}
\label{sec:intersection_processing}

The RIP representation extends the base clustering pipeline by introducing additional geometric preprocessing steps that are applied only when intersection-based features are used. These steps are designed to suppress noisy or spurious intersections while preserving the concentration patterns associated with overlapping plants.

\begin{figure}[h]
    \centering
    \begin{subfigure}[t]{0.48\textwidth}
        \centering
        \includegraphics[width=0.75\textwidth]{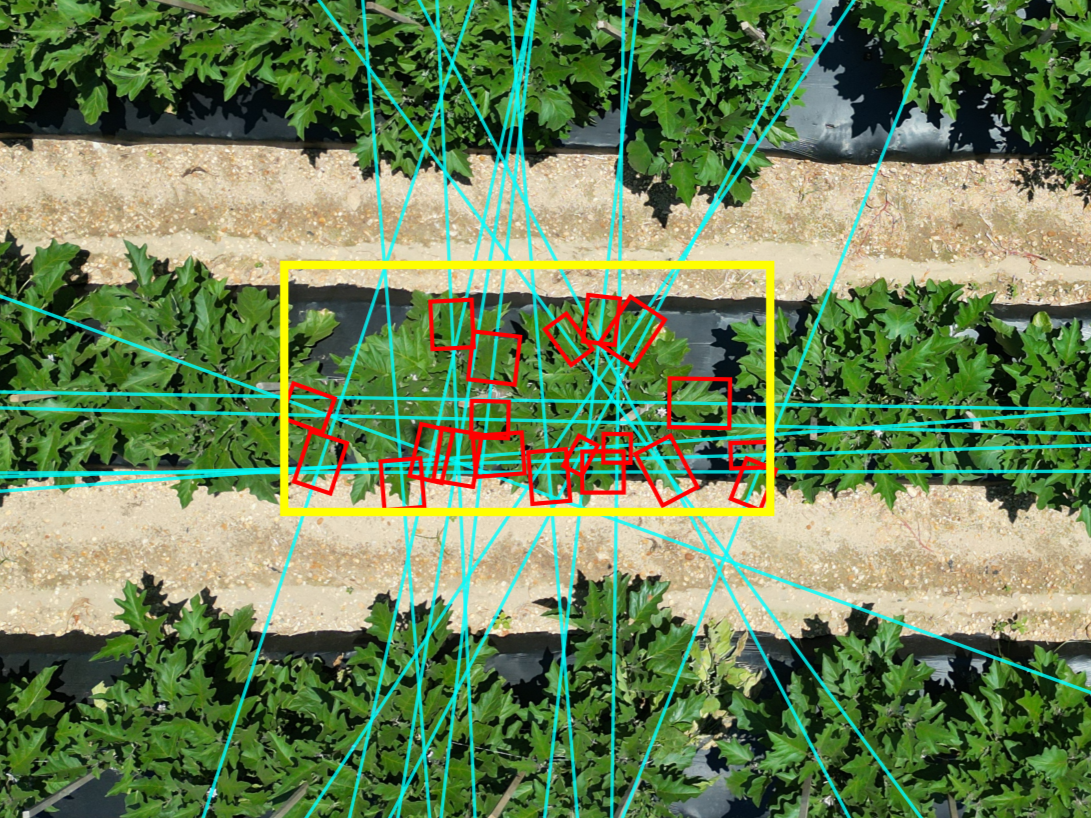}
        \caption{Radial line construction from detected leaf boxes within the bush ROI.}
        \label{fig:radial_line}
    \end{subfigure}
    \hfill
    \begin{subfigure}[t]{0.48\textwidth}
        \centering
        \includegraphics[width=0.75\textwidth]{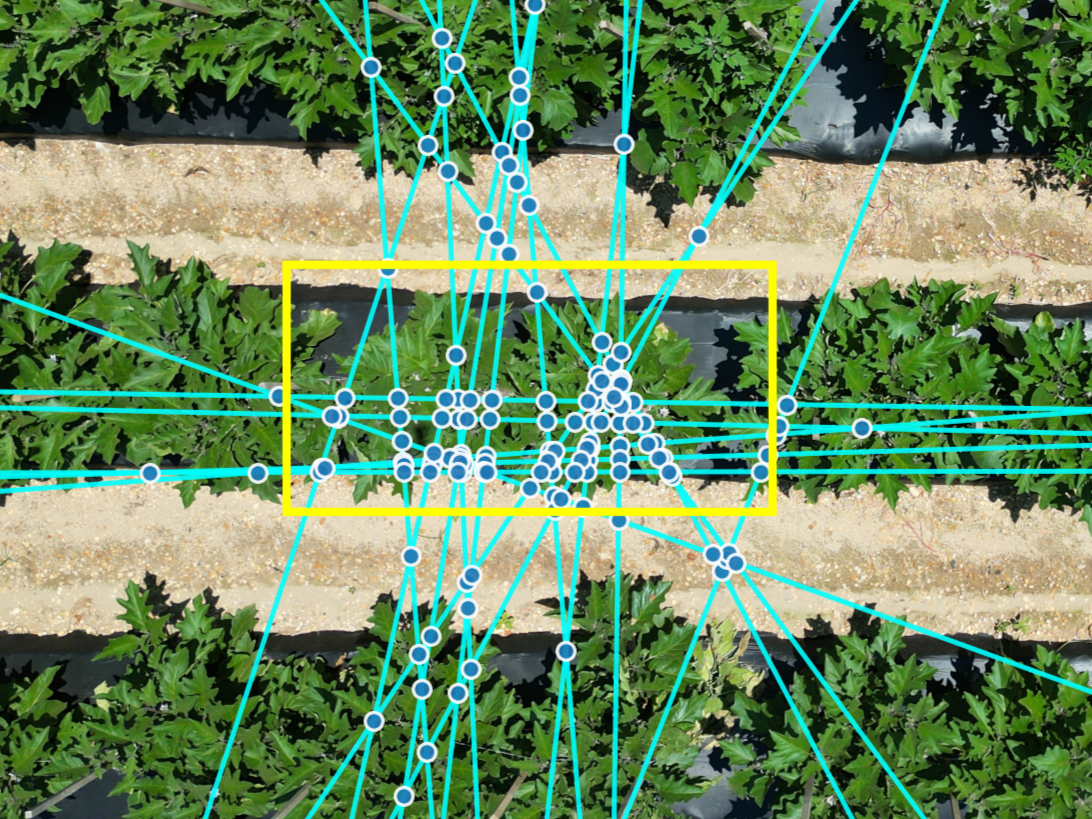}
        \caption{Generation of raw radial intersection points.}
        \label{fig:radial_intersection}
    \end{subfigure}

    \vspace{0.3cm}

    \begin{subfigure}[t]{0.48\textwidth}
        \centering
        \includegraphics[width=0.75\textwidth]{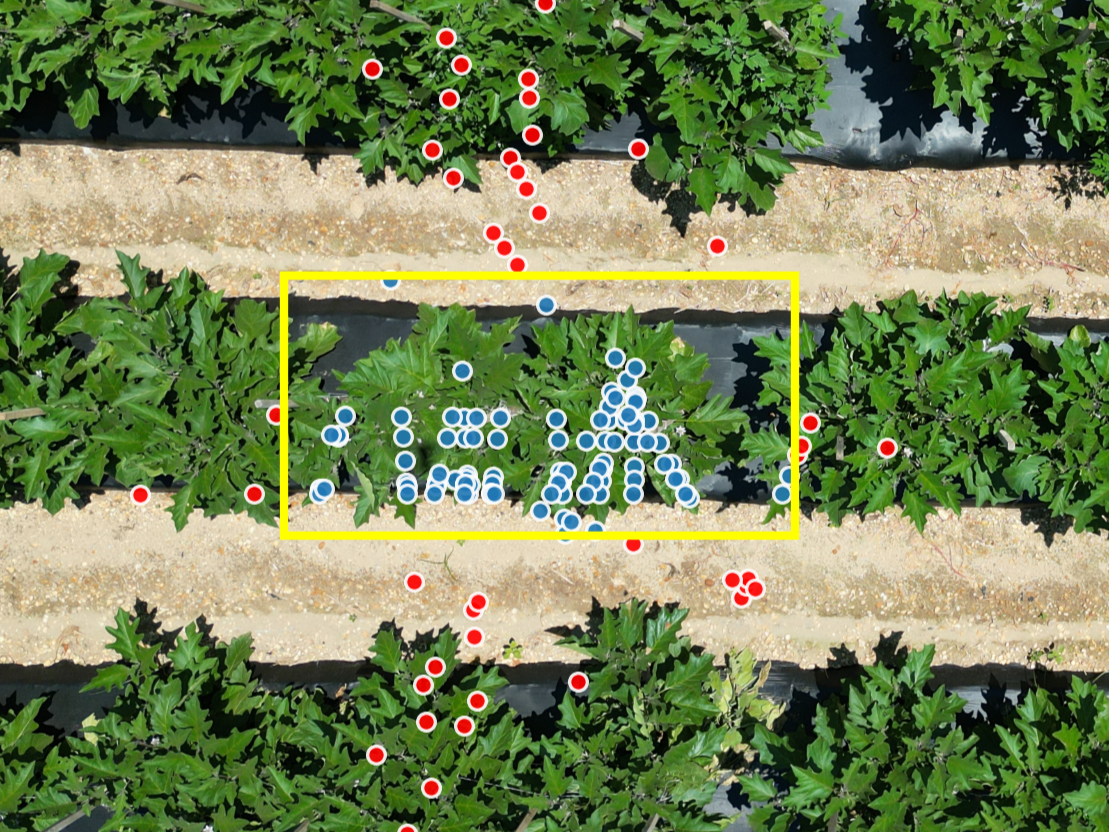}
        \caption{Intersection points restricted to the bush ROI.}
        \label{fig:radial_filtering}
    \end{subfigure}
    \hfill
    \begin{subfigure}[t]{0.48\textwidth}
        \centering
        \includegraphics[width=0.75\textwidth]{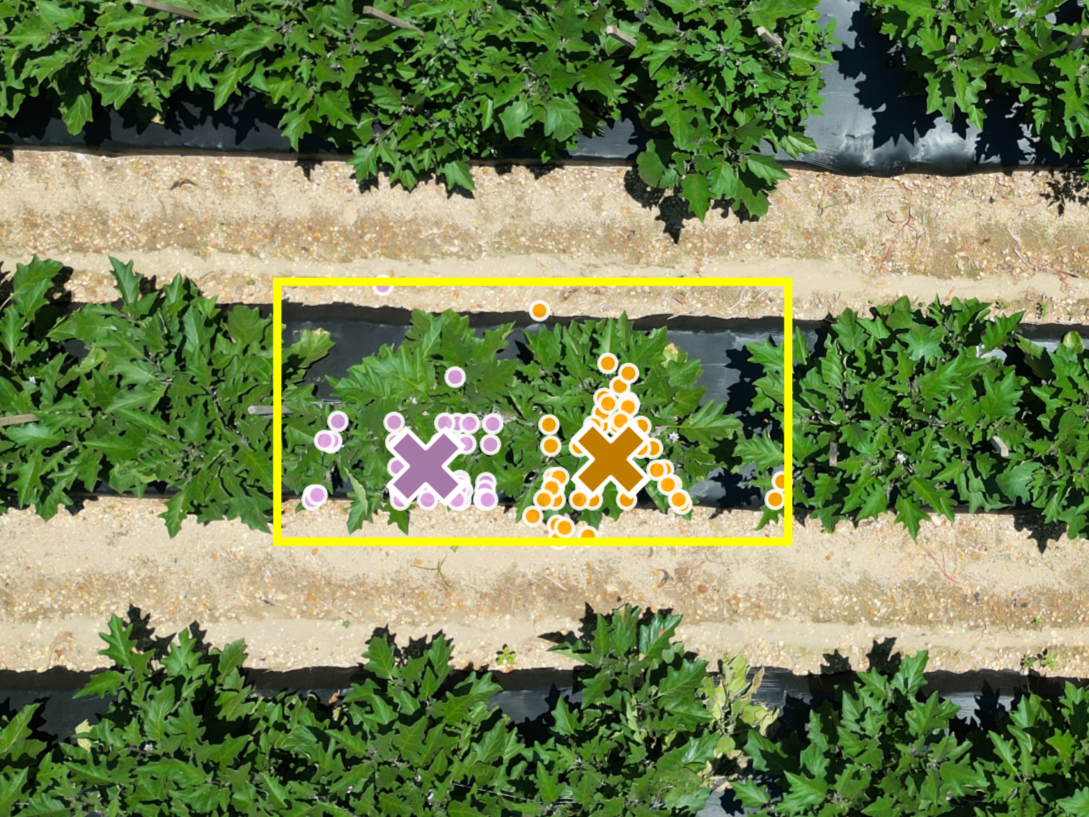}
        \caption{Clustering of retained intersection points.}
        \label{fig:radial_clustering}
    \end{subfigure}

    \caption{Example of the RIP-based processing pipeline on real drone imagery. (a) Radial line construction from detected leaf boxes. (b) Generation of raw intersection points. (c) Filtering of intersection points based on ROI membership. (d) Clustering of retained points into two groups. Cluster centroids are indicated by $\times$ symbols.}
    \label{fig:real_radial_pipeline}
\end{figure}

All intersection-specific operations are performed before the base clustering pipeline described in Section~\ref{sec:base_pipeline}. Once the filtered intersection point set is obtained, clustering and classification proceed exactly as in the centroid-based case. Figure~\ref{fig:real_radial_pipeline} provides a step-by-step visualization of the RIP-based processing pipeline on real drone imagery.

\paragraph{Radial line construction}
Each leaf detection is represented by an oriented bounding box. From this geometry, we compute a unit direction vector aligned with the longest side of the box, representing the dominant leaf orientation. An infinite line is then defined through the leaf centroid in this direction. These lines are interpreted as approximate radial projections of the underlying center of the plant (see Figure \ref{fig:radial_line}).

\paragraph{Intersection point generation}
Given the set of radial lines associated with a bush ROI, we compute all pairwise line--line intersections (see Figure \ref{fig:radial_intersection}). These pairwise intersections define the set of raw intersection points. We then restrict this set to points that lie inside the bush bounding box, yielding the ROI-restricted intersection set (see Figure \ref{fig:radial_filtering}), denoted by $P_{\mathrm{int}}$.

\paragraph{Minimum-density filtering}
The ROI-restricted intersection points may include outliers caused by imperfect detections or orientation noise. To suppress such artifacts, we partition the ROI into a fixed grid and count the number of intersection points in each cell. Points located in cells with density below a threshold $d_{\min}$ are removed. The filtered set is denoted by $P_{\mathrm{int}}^{\ast}$.

\begin{figure}[H]
\setlength{\fboxrule}{2pt} 
\setlength{\fboxsep}{0pt}
    \centering
    \begin{subfigure}[t]{0.48\textwidth}
        \centering
        \includegraphics[width=0.5\textwidth]{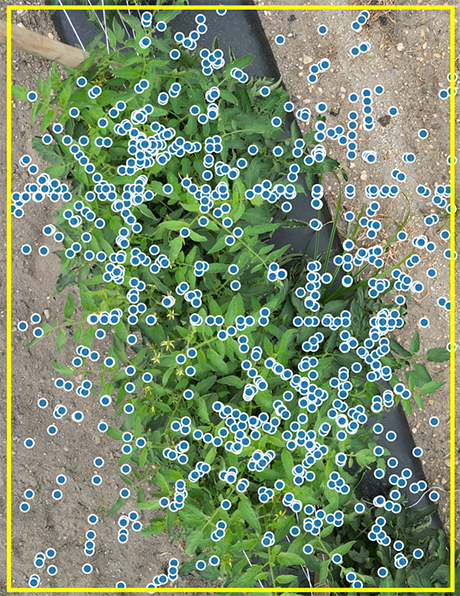}
        \caption{Unfiltered intersection points inside the bush ROI ($P_{\mathrm{int}}$).}
    \end{subfigure}
    \hfill
    \begin{subfigure}[t]{0.48\textwidth}
        \centering
        \fcolorbox{white}{white}{%
        \includegraphics[width=0.5\textwidth]{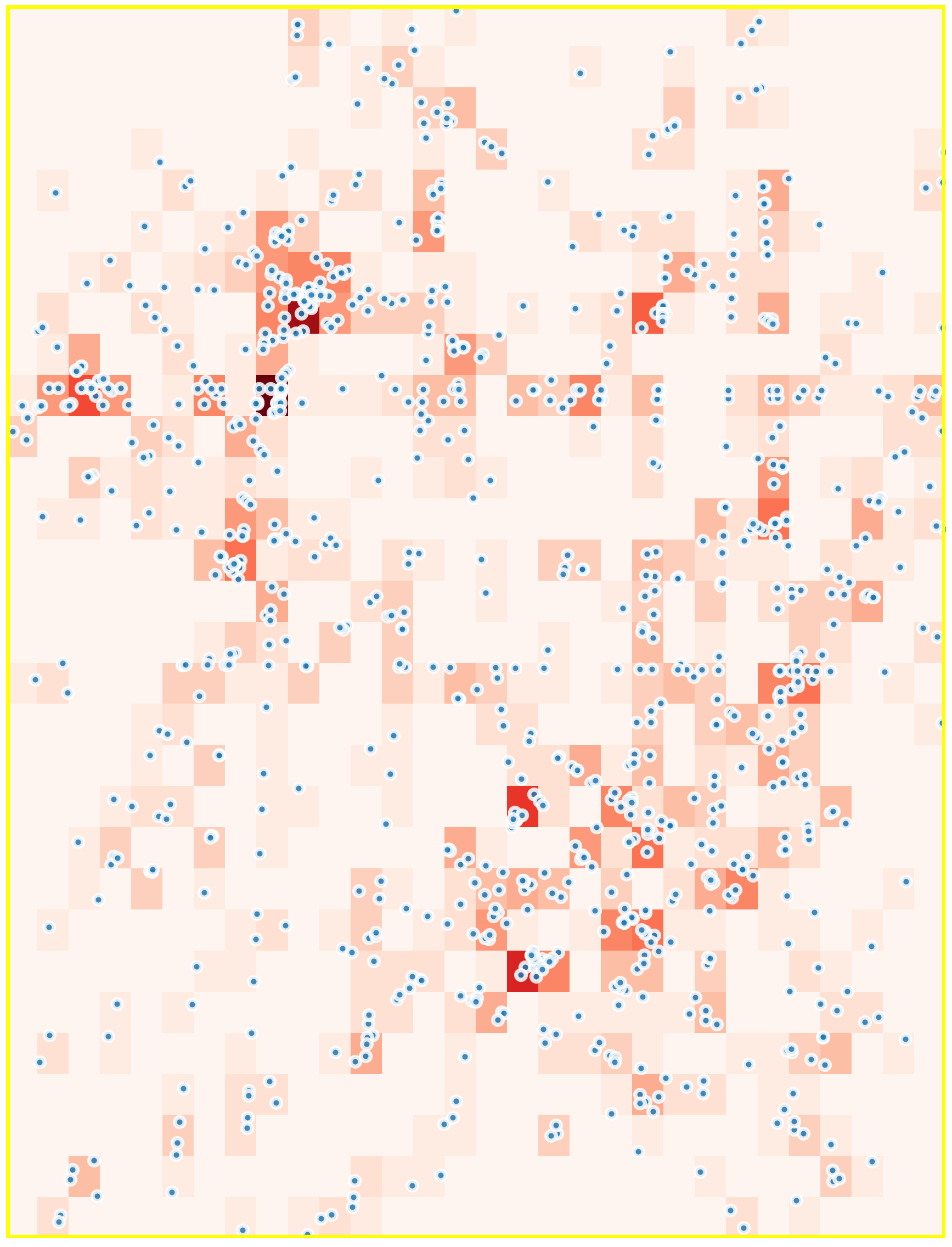}%
        }
        \caption{Density map of $P_{\mathrm{int}}$ (fixed grid, $30 \times 30$ bins).}
    \end{subfigure}

    \vspace{0.3cm}

    \begin{subfigure}[t]{0.48\textwidth}
        \centering
        \fcolorbox{white}{white}{%
        \includegraphics[width=0.5\textwidth]{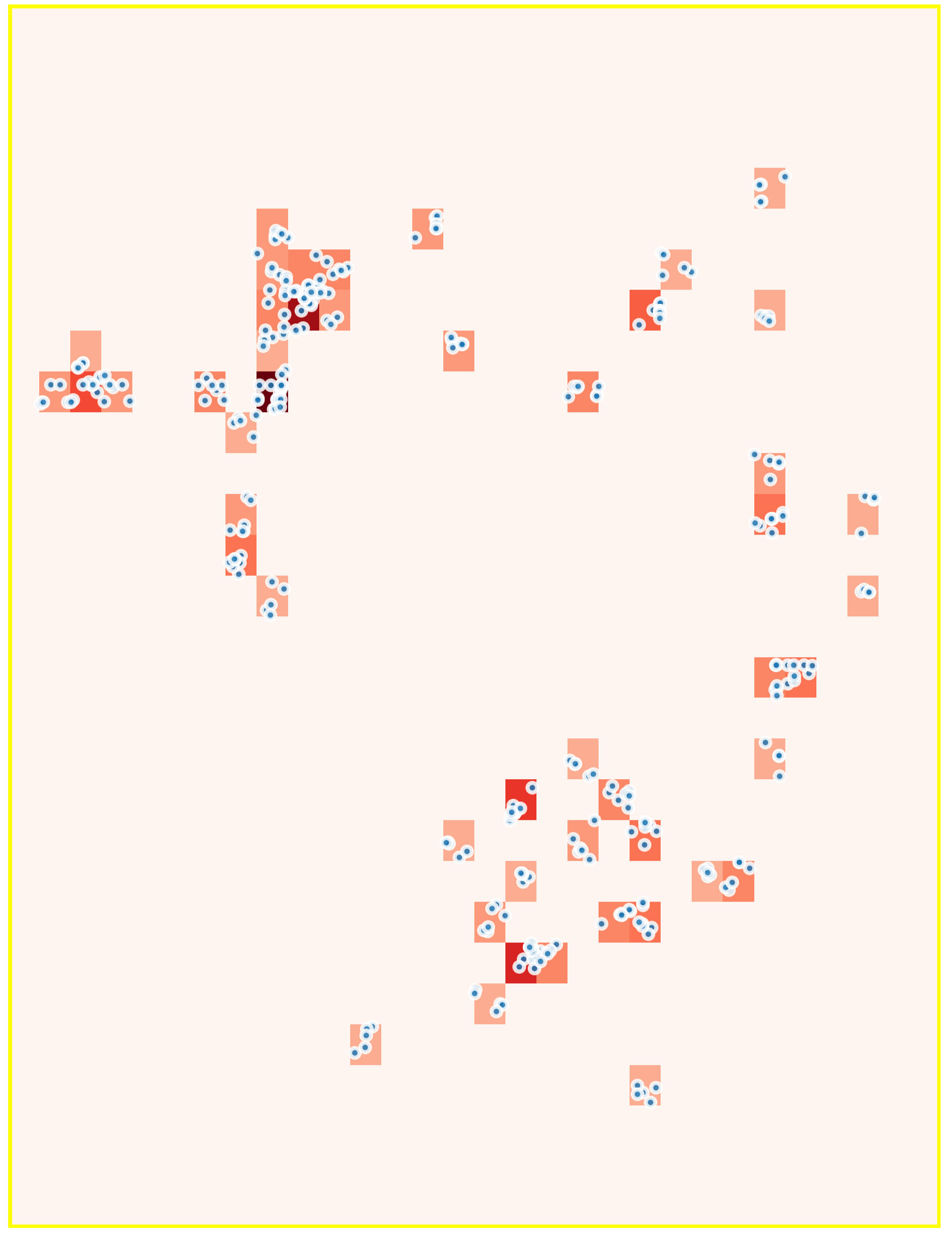}%
        }
        \caption{Density map after minimum-density filtering.}
    \end{subfigure}
    \hfill
    \begin{subfigure}[t]{0.48\textwidth}
        \centering
        \includegraphics[width=0.5\textwidth]{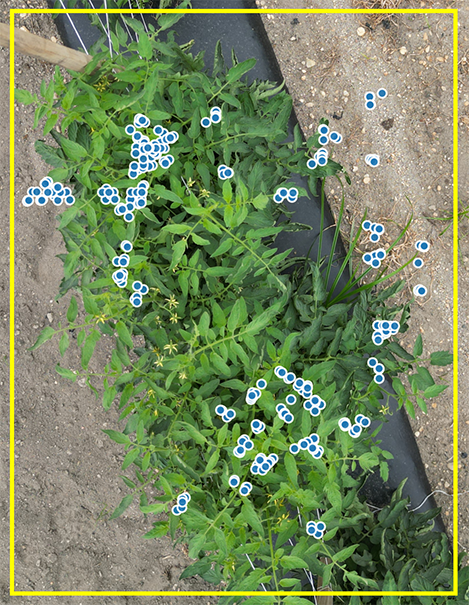}
        \caption{Filtered intersection points ($P_{\mathrm{int}}^{\ast}$).}
    \end{subfigure}

    \caption{Example of minimum-density filtering applied to intersection points. Raw intersection points are first restricted to the ROI and then aggregated into a density map, low-density cells are removed, and the remaining points are retained for downstream clustering.}
    \label{fig:real_min_density_filtering}
\end{figure}

Figure~\ref{fig:real_min_density_filtering} illustrates this process. The ROI-restricted intersection points are first converted into a density map, low-density regions are removed, and the remaining points are used for clustering.

\paragraph{Adaptive density thresholding}
To account for variation in the number of raw intersection points across ROIs, the minimum-density threshold is defined as a function of the total number of intersection points $N = |P_{\mathrm{int}}|$:
\begin{equation}
\label{eq:density_threshold}
\mathrm{min\_density}(N) = \max(1,\; aN + b),
\end{equation}
where the lower bound of 1 prevents the threshold from becoming too small in sparse cases. 
For each clustering backend (K-means and GMM), the coefficients $(a,b)$ are calibrated independently using only the tuning set through grid search over a predefined parameter range, and the pair that maximizes the F1-score for the 2-in-1 class is retained. The selected coefficients are then fixed and applied without further adjustment to the independent validation set. The resulting threshold is therefore an empirically tuned decision rule rather than a fixed theoretical constant.

In practice, grid cells with counts strictly below $\mathrm{min\_density}(N)$ are discarded, and all intersection points falling in those cells are removed.

\paragraph{Safeguard for low-sample cases}
If density filtering removes too many intersection points, the threshold is iteratively relaxed until at least 10 points remain or until the threshold reaches 1. This safeguard ensures that the downstream clustering stage receives a minimally sufficient point set.

\paragraph{Integration with the base pipeline}
The final filtered intersection point set $P_{\mathrm{int}}^{\ast}$ is passed directly to the base clustering pipeline (Section~\ref{sec:base_pipeline}), where two-cluster fitting and silhouette-based classification are performed.

\subsubsection{Ensemble Strategy}
\label{sec:ensemble_strategy}

In addition to evaluating centroid-based and RIP-based representations independently, we introduce a post-pipeline ensemble strategy that combines their predictions at the bush level.

The ensemble operates after both representations have passed through the same base clustering pipeline (Section~\ref{sec:base_pipeline}). Thus, each bush ROI yields:
\begin{itemize}
    \item a centroid-based prediction with silhouette score $s_c$, and
    \item an RIP-based prediction with silhouette score $s_r$.
\end{itemize}
Both predictions are binary (1-in-1 or 2-in-1), obtained using the same silhouette-based decision rule.

\paragraph{Agreement rule}
If both representations produce the same prediction, that prediction is taken as final.

\paragraph{Disagreement resolution}
If the two representations disagree, we select the prediction with higher confidence, defined as the distance of the silhouette score from the decision threshold 0.5:
\[
\mathrm{confidence} = |s - 0.5|.
\]
The prediction associated with the larger confidence value is selected.

This rule favors the representation whose two-cluster structure is more clearly separated for the specific ROI. Centroid-based features capture spatial concentration patterns, whereas RIP-based features capture directional radial structure. The ensemble therefore adaptively selects the representation that better explains the observed geometry.

\paragraph{Rationale}
The centroid-based representation is generally more stable under noisy orientation estimates, whereas the RIP-based representation can better resolve overlapping radial structures when leaf orientations are reliable. The ensemble leverages this complementarity without introducing additional trainable parameters or modifying the clustering formulation.

All ensemble results reported in Section~\ref{sec:results} are computed using this aggregation rule.

\subsubsection{Scope and Limitations}
\label{sec:scope_limitations}

The proposed clustering framework is designed to resolve ambiguous bush-level detections arising from object-detection errors, with a specific focus on distinguishing between single-plant (1-in-1) and overlapping two-plant (2-in-1) cases. Accordingly, several assumptions and limitations apply.

\paragraph{Binary scope and fixed clustering formulation}
This study addresses binary disambiguation between 1-in-1 and 2-in-1 bush-level regions, and all clustering operations are therefore performed with a fixed number of clusters $k = 2$. Detection ROIs containing three or more complete plants were not considered. Extending the framework to higher-order overlaps would require both adapting the clustering formulation to estimate the number of clusters and modifying the decision rule, and is left for future work.

\paragraph{Exclusion of partial bushes}
Bush regions truncated by image boundaries were excluded during dataset construction. Partial visibility can distort both centroid-based and RIP-based geometric representations and therefore complicate interpretation.

\paragraph{Dependence on leaf-level detections}
Clustering performance depends on the quality and spatial coverage of leaf-level detections within each bush ROI. Sparse detections or systematic detection failures can reduce the reliability of both centroid and RIP representations. The ensemble strategy partially mitigates this issue but does not eliminate dependence on upstream detection quality.

\paragraph{Geometric assumptions}
The RIP-based representation assumes leaves or branches have radial geometry. While this assumption is appropriate for the crop types studied here, plants with substantially different growth patterns may require alternative geometric models.

\paragraph{Computational considerations}
RIP construction requires pairwise line--line intersection computations, which can be computationally expensive when the number of leaf detections is very large. In typical field imagery, however, the number of detections per bush remains moderate, and the method can be parallelized across ROIs.

Overall, the proposed framework should be viewed as a post-detection geometric disambiguation tool for resolving common two-plant misdetections, rather than as a general-purpose plant-counting or clustering framework.

\section{Results and Discussion}
\label{sec:results}

\noindent\textbf{Evaluation setup.}\,
We evaluate eggplant and tomato bush-level ROIs on held-out validation sets of 20 samples each (10 labeled 2-in-1 and 10 labeled 1-in-1). Although the dataset size is limited, it is carefully curated to reflect realistic post-detection ambiguity cases encountered in field conditions. Performance is reported using accuracy, precision, recall, and F1-score for the 2-in-1 class.

We compare centroid-based and RIP-based geometric representations, and also evaluate a post-pipeline ensemble strategy combining both. All clustering results are obtained using the two-cluster formulation ($k=2$) described in Section~\ref{sec:clustering}.

Results are presented in a method-centric order consistent with Section~3: centroid-based clustering is first analyzed as a baseline, followed by RIP-based clustering, and finally the combined centroid--intersection decision rule.

\subsection{Centroid-based Clustering Results}
\label{sec:results_centroid}

\subsubsection{Eggplant}

\noindent
For centroid-based clustering, Table~\ref{tab:egg_centroid_raw} shows that centroid geometry provides a strong standalone signal for eggplant.
Both K-means and GMM achieve 0.90 accuracy and an F1-score of 0.89.
Precision for the 2-in-1 class is perfect (1.00), indicating that predicted 2-in-1 cases are highly reliable.

\begin{figure}[t]
    \centering
    \begin{subfigure}[t]{0.48\linewidth}
        \centering
        \includegraphics[width=0.6\linewidth]{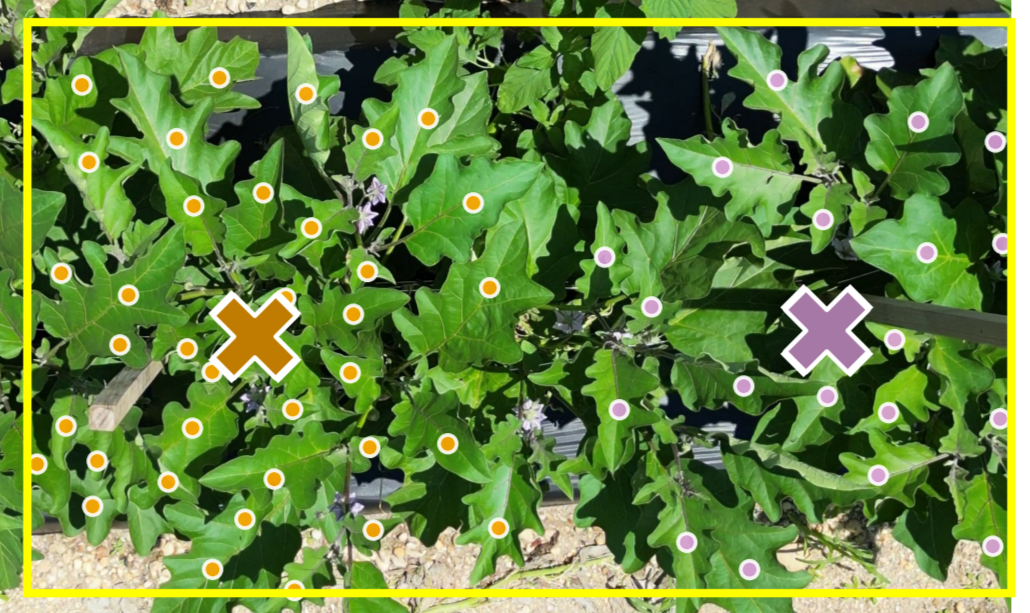}
        \caption{True-positive example. Ground truth: 2-in-1; predicted: 2-in-1. Silhouette score: 0.637.}
        \label{fig:egg_centroid_correct}
    \end{subfigure}
    \hfill
    \begin{subfigure}[t]{0.48\linewidth}
        \centering
        \includegraphics[width=0.55\linewidth]{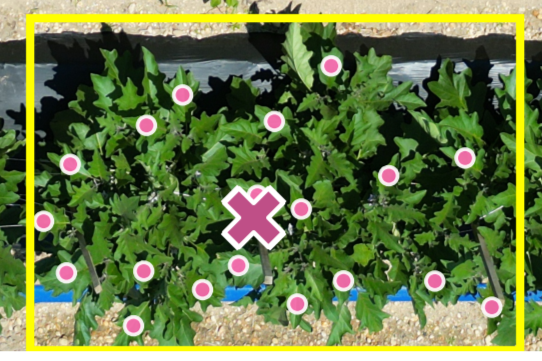}
        \caption{False-negative example. Ground truth: 2-in-1; predicted: 1-in-1. Silhouette score: 0.332.}
        \label{fig:egg_centroid_incorrect}
    \end{subfigure}
    \caption{Representative eggplant centroid-based clustering examples using K-means on ground-truth 2-in-1 ROIs. The left panel shows a correctly resolved case, while the right panel shows a missed 2-in-1 case that is incorrectly classified as 1-in-1 because its silhouette score falls below the decision threshold. Cluster centroids are indicated by $\times$.}
    \label{fig:egg_centroid_examples}
\end{figure}

The remaining errors arise from missed 2-in-1 cases.
In these instances, two partially overlapping point distributions exhibit limited spatial separation between plant centers, resulting in moderate cluster separation.
This is reflected in a silhouette value below the decision threshold (e.g., Figure~\ref{fig:egg_centroid_examples}(b)), leading to incorrect 1-in-1 classification.

\begin{figure*}[h]
    \centering

    \begin{subfigure}[t]{0.48\textwidth}
        \centering
        \includegraphics[width=0.5\linewidth]{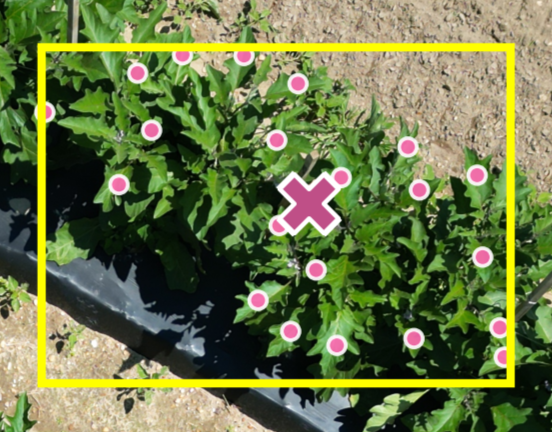}
        \caption{Actual centroid-based pipeline output for a missed 2-in-1 case. The ROI is classified as 1-in-1, with silhouette score $0.471 < 0.5$.}
        \label{fig:egg_centroid_miss1_pred}
    \end{subfigure}
    \hfill
    \begin{subfigure}[t]{0.48\textwidth}
        \centering
        \includegraphics[width=0.5\linewidth]{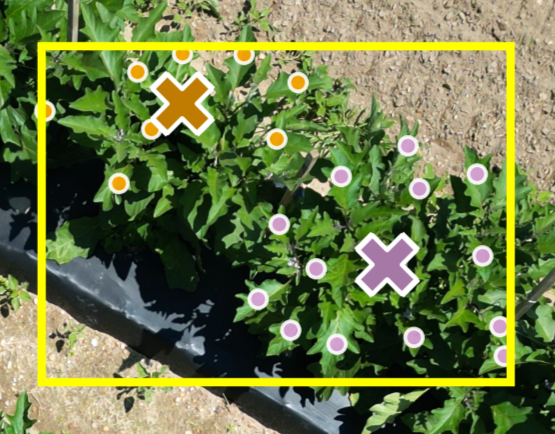}
        \caption{Diagnostic visualization obtained by forcing a two-cluster fit ($k=2$) on the same ROI.}
        \label{fig:egg_centroid_miss1_forced}
    \end{subfigure}

    \vspace{0.4em}

    \caption{Eggplant centroid-based borderline failure case. The left panel shows the actual centroid-based pipeline output for a bush ROI labeled as 2-in-1 but predicted as 1-in-1. The right panel shows the same ROI under a forced two-cluster fit ($k=2$), included for diagnostic comparison. Cluster centroids are marked by $\times$.}
    \label{fig:egg_centroid_failure_forced2}
\end{figure*}

In addition, borderline cases with silhouette scores near the decision threshold (i.e., $s \approx 0.5$) are observed, where clustering structure is weak and classification becomes ambiguous (Figure~\ref{fig:egg_centroid_failure_forced2}).
These cases highlight a limitation of centroid-based geometry under weak spatial separation and illustrate the sensitivity of the decision rule near the threshold.

\begin{figure}[t]
    \centering
    \includegraphics[width=0.225\linewidth]{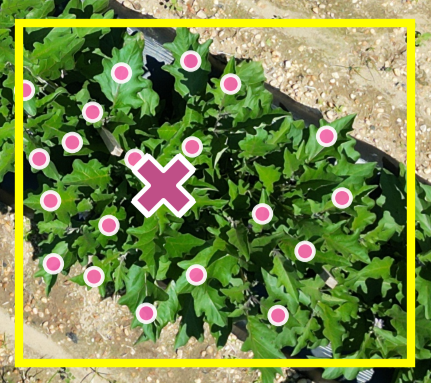}
    \caption{Representative eggplant true-negative example for centroid-based clustering using K-means. The ground-truth label is 1-in-1, and the centroid-based pipeline correctly predicts 1-in-1. Silhouette score: 0.361. Cluster centroid is indicated by $\times$.}
    \label{fig:egg_centroid_tn_example}
\end{figure}

No false-positive 2-in-1 predictions were observed as precision is perfect.
Figure~\ref{fig:egg_centroid_tn_example} shows a representative true-negative case.

\begin{table}[h]
\begin{center}
\begin{tabular}{|c|c|c|}
\hline
Metric & K-means & GMM \\
\hline
Accuracy & 0.90000 & 0.90000 \\
\hline
Precision (2-in-1) & 1.00000 & 1.00000 \\
\hline
Recall (2-in-1) & 0.80000 & 0.80000 \\
\hline
F1-score (2-in-1) & 0.88889 & 0.88889 \\
\hline
\end{tabular}
\caption{Eggplant centroid-based clustering results on the validation set without density filtering.}
\label{tab:egg_centroid_raw}
\end{center}
\end{table}

\noindent\textbf{Centroid takeaway (eggplant).}
For eggplant, centroid geometry alone captures most of the separability between 1-in-1 and 2-in-1 cases and provides a strong baseline.

\subsubsection{Tomato}

\noindent
For tomato, centroid-based clustering achieves moderate performance, with 0.75 accuracy and 0.67 F1-score (Table~\ref{tab:tomato_centroid_based}).
Precision remains 1.00, while recall drops to 0.50, indicating conservative predictions.

\begin{figure}[t]
    \centering
    \begin{subfigure}[t]{0.48\linewidth}
        \centering
        \includegraphics[width=0.5\linewidth]{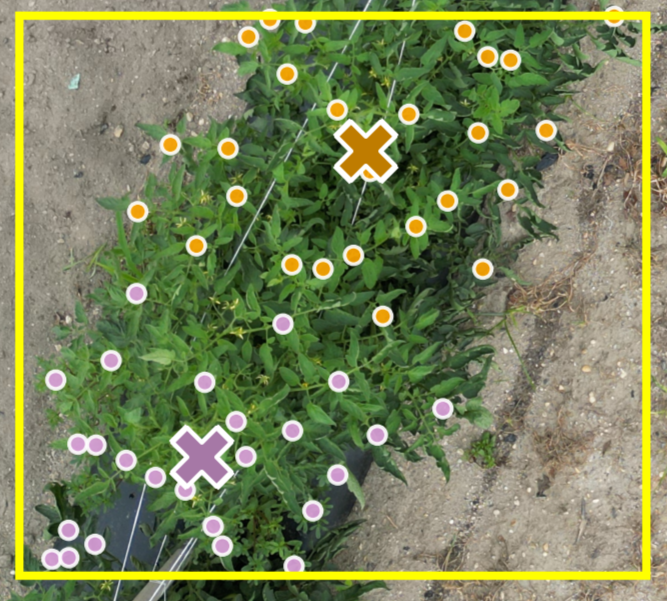}
        \caption{True-positive example. Ground truth: 2-in-1; predicted: 2-in-1. Silhouette score: 0.520.}
        \label{fig:tomato_centroid_correct}
    \end{subfigure}
    \hfill
    \begin{subfigure}[t]{0.48\linewidth}
        \centering
        \includegraphics[width=0.65\linewidth]{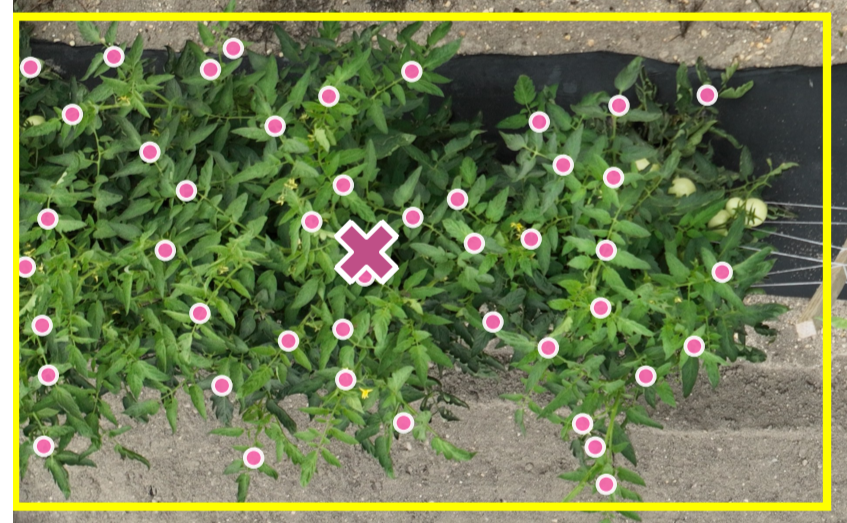}
        \caption{False-negative example. Ground truth: 2-in-1; predicted: 1-in-1. Silhouette score: 0.393.}
        \label{fig:tomato_centroid_incorrect}
    \end{subfigure}
    \caption{Representative tomato centroid-based clustering examples using K-means on ground-truth 2-in-1 ROIs. The left panel shows a correctly resolved case, while the right panel shows a missed 2-in-1 case that is incorrectly classified as 1-in-1 because its silhouette score falls below the decision threshold. Cluster centroids are indicated by $\times$.}
    \label{fig:tomato_centroid_examples}
\end{figure}

In contrast to eggplant, centroid-based clustering is less reliable for tomato.
Overlapping branch structures often reduce cluster separability, leading to missed 2-in-1 detections, as illustrated in Figure~\ref{fig:tomato_centroid_examples}.

Because precision is 1.00, no false-positive 2-in-1 predictions were observed.

\begin{table}[h]
\begin{center}
\begin{tabular}{|c|c|c|}
\hline
Metric & K-means & GMM \\
\hline
Accuracy & 0.75000 & 0.75000 \\
\hline
Precision (2-in-1) & 1.00000 & 1.00000 \\
\hline
Recall (2-in-1) & 0.50000 & 0.50000 \\
\hline
F1-score (2-in-1) & 0.66667 & 0.66667 \\
\hline
\end{tabular}
\caption{Tomato centroid-based clustering results on the validation set without density filtering.}
\label{tab:tomato_centroid_based}
\end{center}
\end{table}

\noindent\textbf{Centroid takeaway (tomato).}
Centroid geometry provides a useful but incomplete signal for tomato, motivating the use of additional geometric cues.

\subsection{RIP-based Clustering Results}
\label{sec:results_rip}

\noindent\textbf{RIP variants.}
We evaluate radial intersection point (RIP)-based features under two configurations: 
(i) without filtering and 
(ii) with minimum-density filtering. 
This comparison isolates the effect of density-based pruning on the quality of the intersection representation.

\subsubsection{Eggplant}

\noindent\textbf{RIP without filtering.}
Without filtering, RIP-based clustering performs worse than centroid-based clustering for eggplant (Table~\ref{tab:egg_int_none}). K-means achieves 0.80 accuracy and an F1-score of 0.78, while GMM achieves 0.75 accuracy and an F1-score of 0.71. This reduction is expected because the RIP representation depends on pairwise geometric interactions among leaf detections and is therefore more sensitive to missed detections and orientation noise.

\begin{figure}[t]
    \centering
    \begin{subfigure}[t]{0.48\linewidth}
        \centering
        \includegraphics[width=0.5\linewidth]{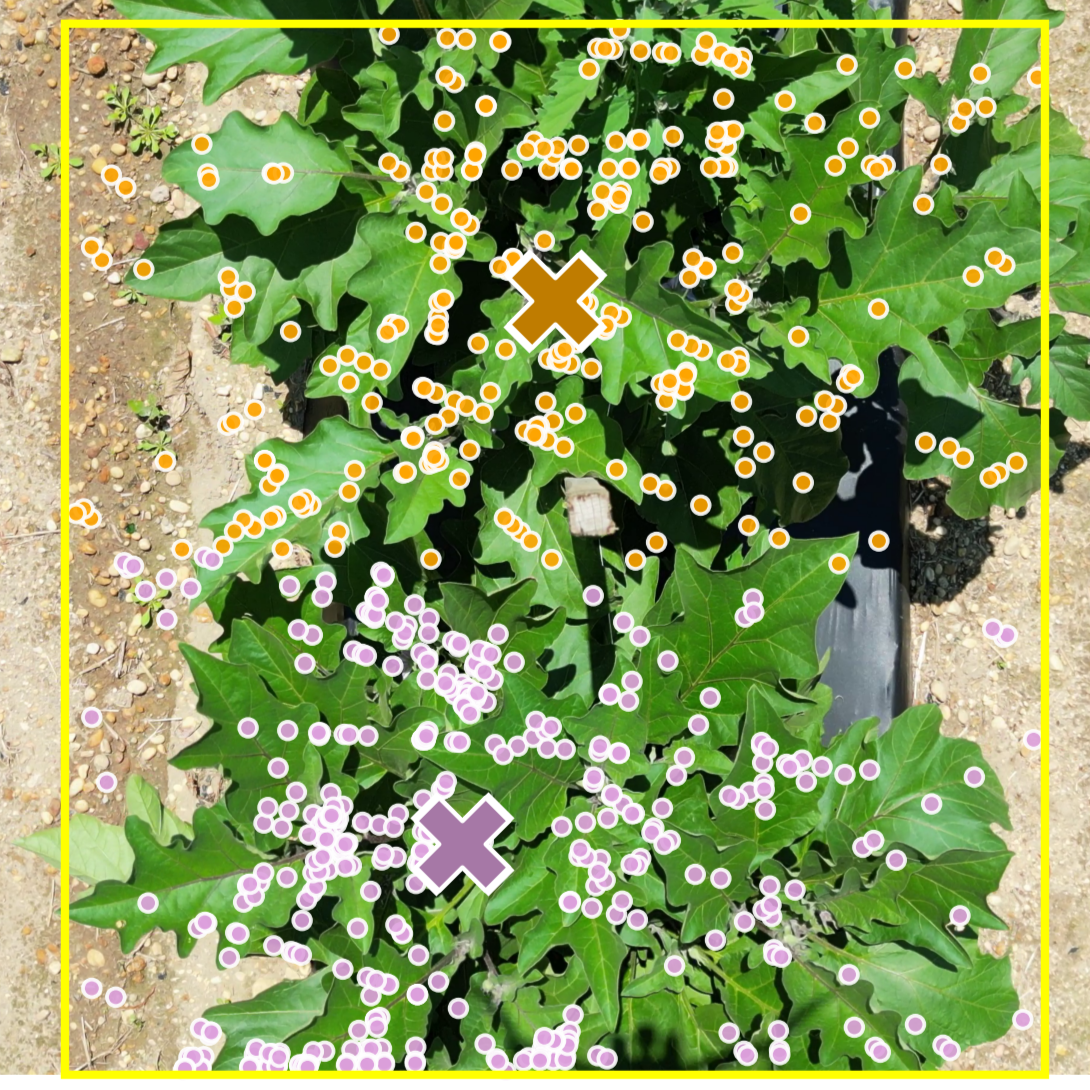}
        \caption{True-positive example. Ground truth: 2-in-1; predicted: 2-in-1. Silhouette score: 0.564.}
        \label{fig:egg_rip_correct}
    \end{subfigure}
    \hfill
    \begin{subfigure}[t]{0.48\linewidth}
        \centering
        \includegraphics[width=0.6\linewidth]{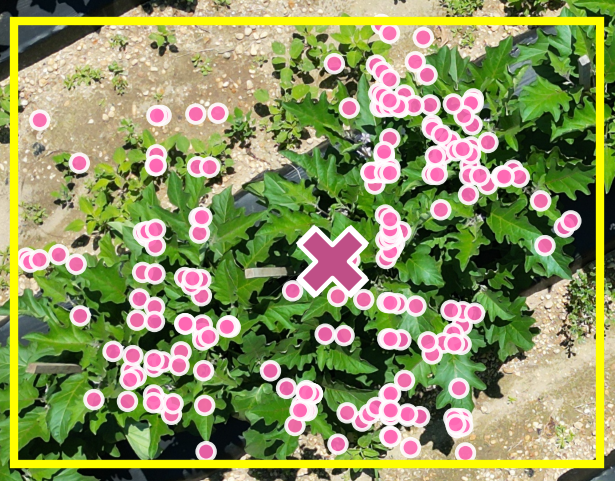}
        \caption{False-negative example. Ground truth: 2-in-1; predicted: 1-in-1. Silhouette score: 0.481.}
        \label{fig:egg_rip_incorrect}
    \end{subfigure}
    \caption{Representative eggplant RIP-based clustering examples without density filtering using K-means on ground-truth 2-in-1 ROIs. The left panel shows a correctly resolved case, while the right panel shows a missed 2-in-1 case that is incorrectly classified as 1-in-1 because its silhouette score falls below the decision threshold. Cluster centroids are indicated by $\times$.}
    \label{fig:egg_rip_examples}
\end{figure}

Figure~\ref{fig:egg_rip_examples} shows representative eggplant examples without density filtering. The left panel illustrates a correctly resolved 2-in-1 case, whereas the right panel shows a missed overlapping case with a silhouette score below the classification threshold.

\begin{table}[h]
\begin{center}
\begin{tabular}{|c|c|c|}
\hline
Metric & K-means & GMM \\
\hline
Accuracy & 0.80000 & 0.75000 \\
\hline
Precision (2-in-1) & 0.87500 & 0.85714 \\
\hline
Recall (2-in-1) & 0.70000 & 0.60000 \\
\hline
F1-score (2-in-1) & 0.77778 & 0.70588 \\
\hline
\end{tabular}
\caption{Eggplant intersection-based clustering results on the validation set using radial intersection points (RIP) without density thresholding.}
\label{tab:egg_int_none}
\end{center}
\end{table}

\noindent\textbf{Effect of minimum-density filtering.}\,
Applying minimum-density filtering modifies the RIP representation by suppressing low-density intersection regions and preserving more concentrated geometric structure.

For eggplant, filtering increases recall for the 2-in-1 class to 0.80 for K-means, at the expense of reduced precision (0.67), resulting in an F1-score of 0.73.
This indicates a tradeoff between recovering missed overlaps and introducing false positives.

The minimum-density thresholds are defined as
\begin{equation}
\mathrm{min\_density}_{\mathrm{KM}}(N) = \max\bigl(1,\; 0.0050N + 1.1000\bigr),
\end{equation}
\begin{equation}
\mathrm{min\_density}_{\mathrm{GMM}}(N) = \max\bigl(1,\; 0.0050N + 1.1300\bigr).
\end{equation}

\begin{figure*}[t]
    \centering
    \begin{subfigure}[t]{\textwidth}
        \centering
        \begin{subfigure}[t]{0.48\textwidth}
            \centering
            \includegraphics[width=0.6\linewidth]{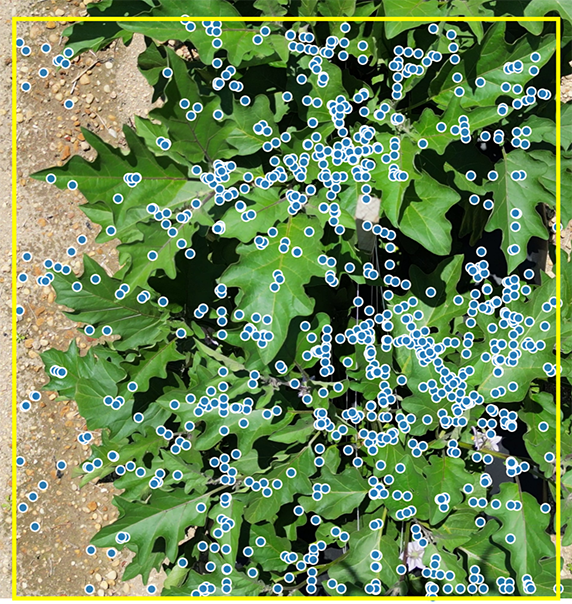}
            \caption{Raw intersection points for a ground-truth 2-in-1 ROI.}
            \label{fig:egg_minden_correct_p1}
        \end{subfigure}
        \hfill
        \begin{subfigure}[t]{0.48\textwidth}
            \centering
            \includegraphics[width=0.6\linewidth]{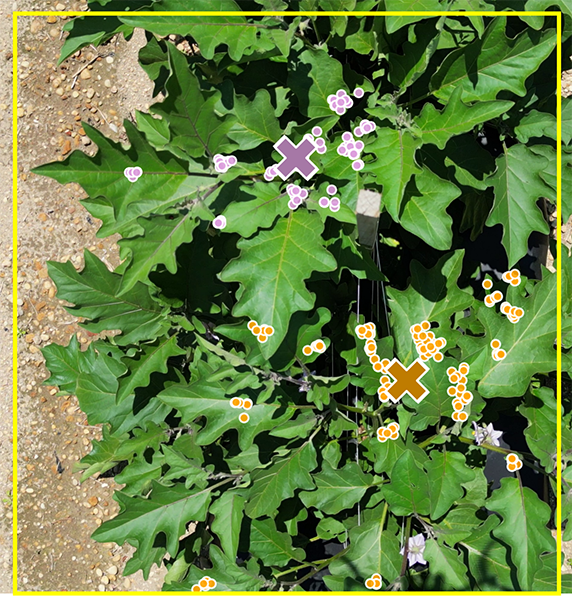}
            \caption{True-positive example. Ground truth: 2-in-1; predicted: 2-in-1. Silhouette score: 0.645.}
            \label{fig:egg_minden_correct_p4}
        \end{subfigure}
    \end{subfigure}

    \vspace{0.6em}

    \begin{subfigure}[t]{\textwidth}
        \centering
        \begin{subfigure}[t]{0.48\textwidth}
            \centering
            \includegraphics[width=0.6\linewidth]{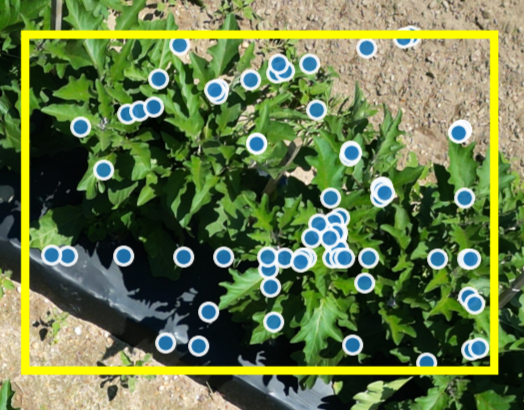}
            \caption{Raw intersection points for a second ground-truth 2-in-1 ROI.}
            \label{fig:egg_minden_wrong_p1}
        \end{subfigure}
        \hfill
        \begin{subfigure}[t]{0.48\textwidth}
            \centering
            \includegraphics[width=0.6\linewidth]{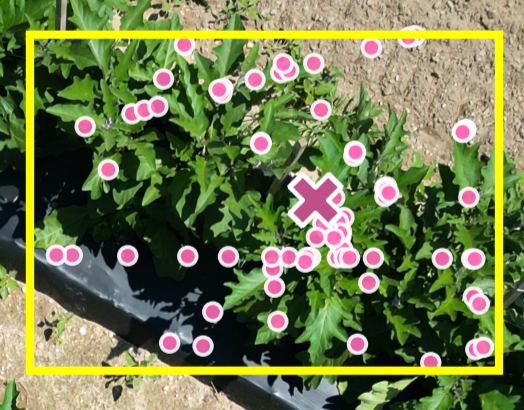}
            \caption{False-negative example. Ground truth: 2-in-1; predicted: 1-in-1. Silhouette score: 0.417.}
            \label{fig:egg_minden_wrong_p4}
        \end{subfigure}
    \end{subfigure}

    \caption{Representative eggplant RIP-based examples obtained with minimum-density filtering and K-means for ground-truth 2-in-1 ROIs. In each row, the left panel shows the raw intersection points and the right panel shows the final clustering result. The top row shows a correctly resolved case, while the bottom row shows a missed 2-in-1 case. Cluster centroids are indicated by $\times$.}
    \label{fig:egg_minden_kmeans_examples}
\end{figure*}

Figure~\ref{fig:egg_minden_kmeans_examples} shows representative eggplant results after minimum-density filtering using K-means. The top row shows a correctly resolved 2-in-1 case, while the bottom row shows a missed 2-in-1 case that remains difficult even after filtering.

\begin{table}[h]
\begin{center}
\begin{tabular}{|c|c|c|}
\hline
Metric & K-means & GMM \\
\hline
Accuracy & 0.70000 & 0.65000 \\
\hline
Precision (2-in-1) & 0.66667 & 0.66667 \\
\hline
Recall (2-in-1) & 0.80000 & 0.60000 \\
\hline
F1-score (2-in-1) & 0.72727 & 0.63158 \\
\hline
\end{tabular}
\caption{Eggplant intersection-based clustering results on the validation set using RIP with density thresholding.}
\label{tab:egg_int_density}
\end{center}
\end{table}

\FloatBarrier

\begin{figure}[t]
    \centering
    \begin{subfigure}[t]{0.48\textwidth}
        \centering
        \includegraphics[width=0.6\linewidth]{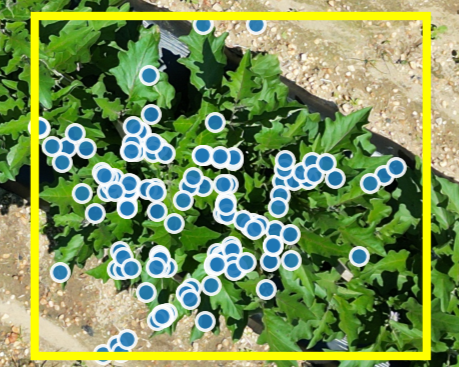}
        \caption{Raw intersection points for a ground-truth 1-in-1 ROI.}
        \label{fig:egg_filtered_rip_false_positive_raw}
    \end{subfigure}
    \hfill
    \begin{subfigure}[t]{0.48\textwidth}
        \centering
        \includegraphics[width=0.6\linewidth]{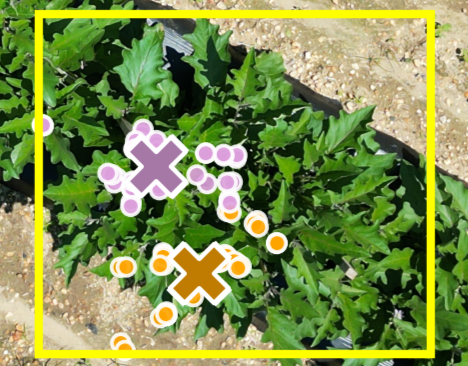}
        \caption{False-positive example after filtering. Ground truth: 1-in-1; predicted: 2-in-1. Silhouette score: 0.519.}
        \label{fig:egg_filtered_rip_false_positive_result}
    \end{subfigure}
    \caption{Representative eggplant false-positive RIP result after minimum-density filtering using K-means. The left panel shows the raw intersection points, and the right panel shows the final filtered clustering output. Although the ROI is labeled as 1-in-1, the filtered RIP pipeline predicts 2-in-1. Cluster centroids are indicated by $\times$.}
    \label{fig:egg_filtered_rip_false_positive_example}
\end{figure}

Figure~\ref{fig:egg_filtered_rip_false_positive_example} illustrates a representative trade-off case after density filtering. In this example, the filtered RIP pipeline produces a false-positive 2-in-1 prediction for a ground-truth 1-in-1 ROI. This behavior is consistent with the drop in precision observed in Table~\ref{tab:egg_int_density} and highlights the fact that density filtering can improve sensitivity while also increasing over-separation in some cases.

\subsubsection{Tomato}

\noindent\textbf{RIP without filtering.}
Without filtering, RIP-based clustering performs poorly for tomato (Table~\ref{tab:tomato_int_none}). K-means achieves 0.55 accuracy and an F1-score of 0.31, while GMM performs even worse on the 2-in-1 class. This reflects the high sensitivity of raw intersection geometry to incomplete or noisy branch detections in tomato.

\begin{table}[h]
\begin{center}
\begin{tabular}{|c|c|c|}
\hline
Metric & K-means & GMM \\
\hline
Accuracy & 0.55000 & 0.55000 \\
\hline
Precision (2-in-1) & 0.66667 & 1.00000 \\
\hline
Recall (2-in-1) & 0.20000 & 0.10000 \\
\hline
F1-score (2-in-1) & 0.30769 & 0.18182 \\
\hline
\end{tabular}
\caption{Tomato intersection-based clustering results on the validation set using radial intersection points (RIP) without density thresholding.}
\label{tab:tomato_int_none}
\end{center}
\end{table}


\noindent\textbf{Effect of minimum-density filtering.}\,
Minimum-density filtering substantially improves RIP-based clustering for tomato.

The thresholds are defined as
\begin{equation}
\mathrm{min\_density}_{\mathrm{KM}}(N) = \max\bigl(1,\; 0.0050N + 0.1500\bigr),
\end{equation}
\begin{equation}
\mathrm{min\_density}_{\mathrm{GMM}}(N) = \max\bigl(1,\; 0.0070N + 2.0000\bigr).
\end{equation}

With filtering, K-means achieves balanced precision and recall (both 0.70), indicating that density-based pruning effectively suppresses spurious intersections.
In contrast, GMM remains less reliable, highlighting the greater robustness of K-means under noisy branch detections.
This behavior may be attributed to the Gaussian assumption underlying GMM, 
which does not align well with the irregular, non-elliptical, and sometimes linear structures observed in the intersection point distributions.

\begin{table}[h]
\begin{center}
\begin{tabular}{|c|c|c|}
\hline
Metric & K-means & GMM \\
\hline
Accuracy & 0.70000 & 0.35000 \\
\hline
Precision (2-in-1) & 0.70000 & 0.38462 \\
\hline
Recall (2-in-1) & 0.70000 & 0.50000 \\
\hline
F1-score (2-in-1) & 0.70000 & 0.43478 \\
\hline
\end{tabular}
\caption{Tomato intersection-based clustering results on the validation set using RIP with density thresholding.}
\label{tab:tomato_int_density}
\end{center}
\end{table}
\FloatBarrier

\noindent\textbf{RIP takeaway.}
Intersection-based geometry is more sensitive to detection noise than centroid geometry, but minimum-density filtering substantially stabilizes the intersection signal, particularly for tomato.

\paragraph{Ablation analysis (effect of density filtering)}
To isolate the contribution of minimum-density filtering, we compare RIP-based clustering with and without filtering while keeping the clustering backend and decision rule unchanged.

For eggplant, filtering increases recall but reduces precision, indicating a tradeoff between recovering missed overlaps and introducing false positives.
For tomato, filtering improves both accuracy and F1-score, demonstrating that density-based pruning effectively stabilizes intersection geometry under noisier detections.

\subsubsection{Sensitivity to grid resolution}
\label{sec:rip_sensitivity}

The minimum-density filtering stage partitions each bush ROI into a fixed $G \times G$ grid before removing low-density cells. To evaluate robustness to this discretization choice, we examined multiple grid resolutions while keeping all other pipeline components unchanged.

Across both crops, performance varies moderately with grid resolution, but the overall qualitative conclusions remain stable. Moderately coarse grids tend to preserve cluster concentration while suppressing scattered intersections, whereas overly fine grids can fragment the density structure and reduce effective support per cell.

\begin{table}[h]
\centering
\begin{tabular}{|c|c|c|c|}
\hline
Crop & Grid & K-means F1 & GMM F1 \\
\hline
 & 25$\times$25 & 0.78 & 0.70 \\
Eggplant & 30$\times$30 & 0.72 & 0.57 \\
 & 35$\times$35 & 0.70 & 0.625 \\
\hline
 & 20$\times$20 & 0.625 & 0.67 \\
Tomato & 25$\times$25 & 0.78 & 0.57 \\
 & 30$\times$30 & 0.70 & 0.43 \\
\hline
\end{tabular}
\caption{Sensitivity of RIP-based clustering performance to grid resolution in the minimum-density filtering stage.}
\label{tab:rip_grid_sensitivity}
\end{table}

Although some configurations produce slightly higher F1-scores, the overall qualitative behavior remains consistent across practical grid sizes. This suggests that the RIP-based pipeline is not overly sensitive to moderate changes in grid resolution.

\subsection{Combined Centroid-RIP Results}
\label{sec:results_combined}

\noindent\textbf{Post-pipeline ensemble.}
We apply the post-pipeline decision rule described in Section~\ref{sec:ensemble_strategy} to combine centroid- and RIP-based predictions.

\subsubsection{Eggplant}

For eggplant, the combined model yields balanced precision and recall of 0.80 for both K-means and GMM (Table~\ref{tab:egg_combined_density}). This improves over some intersection-only variants but remains broadly comparable to the centroid-only baseline, indicating that centroid geometry already captures most of the discriminative structure for this crop.

\begin{table}[h]
\begin{center}
\begin{tabular}{|c|c|c|}
\hline
Metric & K-means & GMM \\
\hline
Accuracy & 0.80000 & 0.80000 \\
\hline
Precision (2-in-1) & 0.80000 & 0.80000 \\
\hline
Recall (2-in-1) & 0.80000 & 0.80000 \\
\hline
F1-score (2-in-1) & 0.80000 & 0.80000 \\
\hline
\end{tabular}
\caption{Eggplant combined centroid-based and intersection-based clustering results on the validation set with density filtering.}
\label{tab:egg_combined_density}
\end{center}
\end{table}
\FloatBarrier

\subsubsection{Tomato}

For tomato, the combined model paired with K-means improves over both centroid-only and RIP-only variants. Accuracy increases to 0.80 with an F1-score of 0.75 and perfect precision for the 2-in-1 class (Table~\ref{tab:tomato_combined_density}). This indicates that density-filtered RIP cues can correct a subset of centroid-based misses. In contrast, the combined GMM result remains weak, reinforcing the greater robustness of K-means for tomato.

\begin{table}[h]
\begin{center}
\begin{tabular}{|c|c|c|}
\hline
Metric & K-means & GMM \\
\hline
Accuracy & 0.80000 & 0.40000 \\
\hline
Precision (2-in-1) & 1.00000 & 0.40000 \\
\hline
Recall (2-in-1) & 0.60000 & 0.40000 \\
\hline
F1-score (2-in-1) & 0.75000 & 0.40000 \\
\hline
\end{tabular}
\caption{Tomato combined centroid-based and intersection-based clustering results on the validation set with density filtering.}
\label{tab:tomato_combined_density}
\end{center}
\end{table}

\subsection{Engineering Implications and Summary of Findings}

The experimental results demonstrate that geometric post-processing can improve the reliability of plant-level interpretation without modifying the underlying object-detection architecture or introducing additional sensing hardware. Centroid-based clustering provides a strong and computationally simple baseline across both crops, whereas RIP-based clustering introduces complementary directional information derived from plant morphology. The effectiveness of these representations depends on crop structure and on the quality of the fine-scale component detections.

For eggplant, centroid-based clustering alone captures most of the discriminative structure between single-plant and overlapping-plant regions. The relatively distinct spatial distribution of detected leaves allows the underlying plant centers to be separated directly from centroid locations. In this case, the additional directional information provided by RIPs does not consistently improve performance and may introduce sensitivity to orientation noise.

Tomato presents a more challenging geometric configuration because branches are more irregular, intertwined, and spatially overlapping. Consequently, centroid separation alone is less reliable. The substantial improvement obtained after density filtering demonstrates that the raw intersection geometry contains useful structural information but requires suppression of spurious intersections before it can be used effectively. The improved performance of the combined centroid--RIP approach with K-means further indicates that spatial and directional geometric cues can provide complementary information when canopy structure becomes more complex.

From an engineering perspective, an important feature of the proposed framework is its modularity. The geometric analysis operates entirely on outputs produced by existing object detectors and therefore does not require retraining or modification of the primary bush detector. It can be implemented as an additional post-processing module in an existing RGB UAV monitoring pipeline, allowing ambiguous bush detections to be further analyzed only when plant-level separation is required. This modular design also permits the upstream detector to be replaced or improved independently of the proposed geometric reasoning stage.

The results also identify a practical trade-off between computational simplicity and structural information. Centroid-based features are inexpensive to compute and relatively robust to orientation errors, making them suitable as a first-stage geometric representation. RIP-based features provide richer directional information but require pairwise line-intersection computations and are more sensitive to errors in fine-scale component orientation. Density filtering reduces this sensitivity by retaining concentrated intersection structures that are more likely to correspond to underlying plant centers.

These observations suggest a practical processing strategy in which centroid geometry provides a robust baseline and directional RIP information is used as a complementary cue for geometrically ambiguous cases. The proposed ensemble implements this principle without introducing additional trainable parameters. More broadly, the results demonstrate that geometric reasoning can recover useful plant-level structural information from standard two-dimensional UAV imagery and can therefore complement deep-learning-based detection without requiring LiDAR, depth sensing, or three-dimensional reconstruction.

Overall, the proposed framework provides a lightweight engineering solution for improving plant-level measurements in dense crop canopies. Its RGB-only formulation, modular architecture, and compatibility with existing object-detection pipelines make it suitable for integration into UAV-based crop-monitoring systems where additional sensing hardware or computationally intensive reconstruction may not be practical.

\FloatBarrier
\section{Conclusion and Future Work}\label{sec5}

This study developed a geometry-aware post-detection framework for resolving overlapping plant instances in RGB UAV imagery. Rather than modifying the primary object detector or introducing additional sensing modalities, the proposed approach extracts plant-level structural information from fine-scale component detections using two complementary geometric representations: component centroids and radial intersection points (RIPs).

Experiments on eggplant and tomato demonstrated that the effectiveness of these representations depends on crop morphology and detection quality. Centroid-based clustering provided the strongest standalone performance for eggplant, where the spatial distribution of detected leaves was sufficient to distinguish most overlapping-plant cases. For tomato, where branch structures were more irregular and intertwined, density filtering substantially improved the robustness of RIP-based clustering. Combining centroid and density-filtered RIP information with K-means provided the best overall tomato result, demonstrating that spatial and directional geometric cues can provide complementary information under more complex canopy conditions.

The main engineering value of the proposed framework lies in its modularity and compatibility with existing UAV-based crop-monitoring systems. Because the geometric analysis operates entirely on outputs from existing object detectors and standard two-dimensional RGB imagery, it can be incorporated as a post-processing module without retraining the primary bush detector or requiring LiDAR, depth sensing, or three-dimensional reconstruction. The framework therefore provides a lightweight approach for improving plant-level interpretation when overlapping canopies produce ambiguous detector outputs.

Several limitations should be addressed before broader operational deployment. The present study considers only single-plant and two-plant overlap cases and excludes partially visible bushes. Future work will extend the framework to estimate an unknown number of plant instances, accommodate partially visible canopies, and evaluate performance on larger datasets across additional crop types, growth stages, and field conditions. Improving the robustness of fine-scale leaf and branch detection will also be important because the geometric representations depend directly on the quality and spatial coverage of these detections.

Future development will further investigate adaptive selection or weighting of centroid and RIP features, segmentation-assisted feature extraction, and learning-assisted geometric filtering. Multi-view or three-dimensional information may also be incorporated when available while retaining the RGB-only approach as a practical baseline. Finally, computational optimization and real-time or near-real-time implementation will be investigated to assess the scalability of the framework for operational UAV-based crop monitoring.

\section{Acknowledgments}

This work was partially funded by Grant \#80NSSC24K0836 awarded to Rowan University (PI: Meenar) by the National Aeronautics and Space Administration (NASA) through the Research Initiation Awards (RIA) program. The authors also thank the managers of Rambo and Tuba Farms, as well as Geography, Planning, and GIS student Graham Luther, for their assistance with data collection and analysis. The authors are solely responsible for the content and any opinions expressed in this paper.

\section*{Data Availability}

The data that support the findings of this study are available from the corresponding author upon reasonable request.

\section*{Declaration of generative AI and AI-assisted technologies in the manuscript preparation process}

During the preparation of this work, the authors used ChatGPT (OpenAI) to assist with language and readability and to generate the illustrative images used in Figure 2. The authors reviewed and edited the output as needed, verified the content for accuracy, and take full responsibility for the content of the publication.


\begin{thebibliography}{99}

\bibitem{Sadashivan2021}
S.~Sadashivan, S.S.~Bhattacherjee, G.~Priyanka, R.~Pachamuthu, J.~Kholova,
Fully automated region of interest segmentation pipeline for UAV based RGB images,
\emph{Biosystems Engineering} 211 (2021) 192--204.
\url{https://doi.org/10.1016/j.biosystemseng.2021.08.032}.

\bibitem{Lee2023Broccoli}
C.-J.~Lee, M.-D.~Yang, H.-H.~Tseng, Y.-C.~Hsu, Y.~Sung, W.-L.~Chen,
Single-plant broccoli growth monitoring using deep learning with UAV imagery,
\emph{Computers and Electronics in Agriculture} 207 (2023) 107739.
\url{https://doi.org/10.1016/j.compag.2023.107739}.

\bibitem{Jiang2024Potato}
H.~Jiang, B.G.~Murengami, L.~Jiang, C.~Chen, C.~Johnson,
F.~Auat Cheein, S.~Fountas, R.~Li, L.~Fu,
Automated segmentation of individual leafy potato stems after canopy consolidation
using YOLOv8x with spatial and spectral features for UAV-based dense crop identification,
\emph{Computers and Electronics in Agriculture} 219 (2024) 108795.
\url{https://doi.org/10.1016/j.compag.2024.108795}.

\bibitem{Blok2021}
P.M.~Blok, E.J.~van Henten, F.K.~van Evert, G.~Kootstra,
Image-based size estimation of broccoli heads under varying degrees of occlusion,
\emph{Biosystems Engineering} 208 (2021) 213--233.
\url{https://doi.org/10.1016/j.biosystemseng.2021.06.001}.

\bibitem{Mirbod2023}
O.~Mirbod, D.~Choi, P.H.~Heinemann, R.P.~Marini, L.~He,
On-tree apple fruit size estimation using stereo vision with deep learning-based
occlusion handling,
\emph{Biosystems Engineering} 226 (2023) 27--42.
\url{https://doi.org/10.1016/j.biosystemseng.2022.12.008}.

\bibitem{Shi2024}
H.~Shi, J.~Zhang, A.~Lei, C.~Wang, Y.~Xiao, C.~Wu, Q.~Wu,
S.~Zhang, J.~Xie,
Enhancing detection accuracy of highly overlapping targets in agricultural imagery
using IoA-SoftNMS algorithm across diverse image sizes,
\emph{Computers and Electronics in Agriculture} 227 (2024) 109475.
\url{https://doi.org/10.1016/j.compag.2024.109475}.

\bibitem{Burusa2024}
A.K.~Burusa, J.~Scholten, X.~Wang, D.~Rapado-Rinc\'on,
E.J.~van Henten, G.~Kootstra,
Semantics-aware next-best-view planning for efficient search and detection
of task-relevant plant parts,
\emph{Biosystems Engineering} 248 (2024) 1--14.
\url{https://doi.org/10.1016/j.biosystemseng.2024.09.018}.

\bibitem{Redmon2016}
J.~Redmon, S.~Divvala, R.~Girshick, A.~Farhadi,
You only look once: Unified, real-time object detection,
in: \emph{Proceedings of the IEEE Conference on Computer Vision and Pattern Recognition (CVPR)},
2016, pp.~779--788.

\bibitem{Ultralytics}
Ultralytics,
YOLOv8 documentation.
\url{https://docs.ultralytics.com/models/yolov8/}.

\bibitem{Kaufman2009}
L.~Kaufman, P.J.~Rousseeuw,
\emph{Finding Groups in Data: An Introduction to Cluster Analysis},
Wiley, 2009.

\end{thebibliography}

\end{document}